\pdfoutput=1

\documentclass[journal]{IEEEtran}

\usepackage[utf8]{inputenc}
\usepackage{graphicx}
\usepackage{paralist}
\usepackage{hyperref}
\usepackage{balance}
\usepackage{amssymb}
\usepackage{multirow}
\usepackage[backend=bibtex,natbib=true,url=false]{biblatex}
\ifCLASSINFOpdf
\else
\fi
\ifCLASSOPTIONcompsoc
    \usepackage[caption=false, font=normalsize, labelfont=sf, textfont=sf]{subfig}
\else
	\usepackage[caption=false, font=footnotesize]{subfig}
\fi

\usepackage{amsthm}
\usepackage{amsmath}
\usepackage{glossaries}
\usepackage[table]{xcolor}
\glsdisablehyper
\newacronym{apaka}{apaka}{automated packaging for autoproj}
\newacronym{cg}{CG}{Causal Graph}
\newacronym{dgps}{DGPS}{Differential Global Positioning System}
\newacronym[firstplural=Degrees of Freedom (DoFs)]{dof}{DoF}{Degree of Freedom}
\newacronym{dtg}{DTG}{Domain Transition Graph}
\newacronym{IMS}{IMS}{Inertial Measurement System}
\newacronym{iBLOCK}{iBLOCK}{intelligent Building Block}
\newacronym{iSSI}{iSSI}{intelligent Space System Interface}
\newacronym{ins}{INS}{Inertial Navigation System}
\newacronym{lrf}{LRF}{Laser Range Finder}
\newacronym{mems}{MEMS}{Micro-Electromechanical System}
\newacronym{mlsm}{MLS\ map}{Multi-Level Surface Map}
\newacronym{mrt}{MRT}{Microsoft Robotics Toolkit}
\newacronym{str}{str}{spatio-temporal requirement}
\newacronym{stqe}{stqe}{spatio-temporally qualified expression}
\newacronym{usar}{USAR}{Urban Search and Rescue}
\newacronym{xml}{XML}{Extensible Markup Language}
\newacronym{ABox}{ABox}{assertional box}
\newacronym{ACC}{ACC}{Agent Communication Channel}
\newacronym{ADL}{ADL}{Action Description Language}
\newacronym{ACL}{ACL}{Agent Communication Language}
\newacronym{ARM}{ARM}{Application Reference Model}
\newacronym{ARP}{ARP}{Address Resolution Protocol}
\newacronym{CG}{CG}{Causal graph}
\newacronym{CCSDS}{CCSDS}{Consultative Committee for Space Data Systems}
\newacronym{CRAM}{CRAM}{Cognitive Robot Abstract Machine}
\newacronym{CORBA}{CORBA}{Common Open Request Broker Architecture}
\newacronym{CPL}{CPL}{CRAM Plan Language}
\newacronym{CVRP}{CVRP}{Capacitated VRP}
\newacronym{CSP}{CSP}{constraint-satisfaction problem}
\newacronym{DARP}{DARP}{Dial-a-Ride Problem}
\newacronym{DGPS}{DGPS}{Differential Global Positioning System}
\newacronym{DL}{DL}{Description Logic}
\newacronym{DDS}{DDS}{Data Distribution Service}
\newacronym{DNS}{DNS}{Domain Name System}
\newacronym{DSD}{DSD}{Distributed Service Directory Service}
\newacronym{DSN}{DSN}{Deep Space Network}
\newacronym{DSL}{DSL}{Domain-specific Language}
\newacronym{DTG}{DTG}{Domain Transition Graph}
\newacronym{ECSS}{ECSS}{European Collaboration for Space Standardization}
\newacronym{EMI}{EMI}{electromechanical interface}
\newacronym{FIPA}{FIPA}{Foundation for Intelligent Physical Agents}
\newacronym{FOL}{FOL}{First Order Language}
\newacronym{FRM}{FRM}{Functional Reference Model}
\newacronym{GLPK}{GLPK}{GNU Linear Programming Kit}
\newacronym{Gecode}{Gecode}{Generic constraint programming framework}
\newacronym{GQR}{GQR}{Generic Qualitative Reasoner}
\newacronym{GPS}{GPS}{Global Positioning System}
\newacronym{HTN}{HTN}{Hierarchical Task Network}
\newacronym{HFVRP}{HFVRP}{Heterogeneous or mixed Fleet VRP}
\newacronym{HWMP}{HWMP}{Hybrid Wireless Mesh Protocol}
\newacronym{iBOSS}{iBOSS}{intelligent Building Block for On-Orbit Satellite Servicing}
\newacronym{IA}{IA}{interval algebra}
\newacronym{IoT}{IoT}{Internet of Things}
\newacronym{ILP}{ILP}{Integer Linear Programming}
\newacronym{IP}{IP}{Internet Protocol}
\newacronym{ISEG}{ISEG}{International Space Exploration Coordination Group}
\newacronym{LP}{LP}{Linear Program}
\newacronym{LDSB}{LDSB}{Lightweight Dynamic Symmetry Breaking}
\newacronym{JADE}{JADE}{Java Agent DEvelopment Framework}
\newacronym{JVM}{JVM}{Java Virtual Machine}
\newacronym{KIF}{KIF}{Knowledge Integration Framework}
\newacronym{MIP}{MIP}{Mixed Integer Progamming}
\newacronym{MTS}{MTS}{message-transport service}
\newacronym{MAS}{MAS}{multi-agent system}
\newacronym{MOISE+}{MOISE+}{Model of Organisation for multI-agent SystEms}
\newacronym{MoreOrg}{MoreOrg}{Model for Reconfigurable Multi-Robot Organisations}
\newacronym{MRS}{MRS}{multi-robot system}
\newacronym{MTVRP}{MTVRP}{Multi-Trip VRP}
\newacronym{NAT}{NAT}{Network Address Translation}
\newacronym{OMNI}{OMNI}{Organisational Model for Normative Institutions}
\newacronym{OMACS}{OMACS}{Organisation Model for Adaptive Computational Systems}
\newacronym{ORA}{ORA}{Ontologies for Robotics and Automation}
\newacronym{OSI}{OSI}{Open System Interconnect}
\newacronym{RASDS}{RASDS}{Reference Architecture for Space Data Systems}
\newacronym{RBox}{RBox}{role box}
\newacronym{Rock}{Rock}{Robot Construction Kit}
\newacronym{ROS}{ROS}{Robot Operating System}
\newacronym{RMRS}{RMRS}{reconfigurable multi-robot system}
\newacronym{PA}{PA}{point algebra}
\newacronym{PPVRP}{PPVRP}{Pallet Packing VRP}
\newacronym{PDDL}{PDDL}{Planning-Domain Definition Language}
\newacronym{PDP}{PDP}{Pickup Delivery Problem}
\newacronym{SLAM}{SLAM}{simultaneous localization and mapping}
\newacronym{SDS}{SDS}{Service Directory Service}
\newacronym{STN}{STN}{simple temporal network}
\newacronym{SWRL}{SWRL}{Semantic Web Rule Language}
\newacronym{TemPl}{TemPl}{Temporal Planning for Reconfigurable Systems}
\newacronym{TCP}{TCP}{Transmission Control Protocol}
\newacronym{TBox}{TBox}{terminological box}
\newacronym{TCN}{TCN}{temporal constraint network}
\newacronym{TCSP}{TCSP}{temporal constraint satisfaction problem}
\newacronym{Templ}{TemPl}{Temporal Planning for Reconfigurable Multi-Robot Systems}
\newacronym{TSP}{TSP}{Travelling Salesman Problem}
\newacronym{TTRP}{TTRP}{Truck-and-Trailer Routing Problem}
\newacronym{VRPTT}{VRPTT}{VRP with Trailers and Transshipments}
\newacronym{VRPMSs}{VRPMSs}{VRP with multiple synchronization constraints}
\newacronym{MMCF}{MMCF}{multi-commodity min-cost flow problem}
\newacronym{ORM}{ORM}{Operations Reference Model}
\newacronym{OWL}{OWL}{Web Ontology Language}
\newacronym{OWL2}{OWL 2}{Web Ontology Language 2}
\newacronym{RDF}{RDF}{Resource Description Framework}
\newacronym{SIMA}{SIMA}{Symmetrical Interface Manipulator}
\newacronym{W3C}{W3C}{World Wide Web Consortium}
\newacronym{VPN}{VPN}{Virtual Private Network}
\newacronym{VRP}{VRP}{Vehicle Routing Problem}
\newacronym{DCVRP}{DCVRP}{Distance-constrained Capacitated VRP}
\newacronym{UDP}{UDP}{User Datagram Protocol}
\newacronym{UDT}{UDT}{UDP-based Data Transfer Protocol}
\newacronym{URDF}{URDF}{Universal Robot Description Format}
\newacronym{VRPTW}{VRPTW}{VRP with Time Windows}
\newacronym{VRPSTW}{VRPSTW}{VRP with Soft Time Windows}
\newacronym{VRPC}{VRPC}{VRP with Compartments}
\newacronym{VRPM}{VRPM}{VRP with Multiple use of vehicles}
\newacronym{VRPPD}{VRPPD}{VRP with Pick-up and Delivery}
\newacronym{VLNS}{VLNS}{Very Large Neighbourhood Search}
\newacronym{MDHFVRPTW}{MDHFVRPTW}{Multi-depot heterogeneous fleet VRP with time windows}
\newacronym{HVRPFD}{HVRPFD}{Heterogeneous VRP with Fixed Costs and Vehicle Dependent Routing Costs}
\newacronym{2L-CVRP}{2L-CVRP}{CVRP with 2-dimensional Loading Constraints}

\makeatletter
\newlength{\limitedheadlinewidth}
\newlength{\limitedlinewidth}
\newtheoremstyle{2maz_definition}
  {6pt}% space before
  {3pt}% space after
  {
    \setlength{\limitedheadlinewidth}{\linewidth}
    \addtolength{\limitedheadlinewidth}{-.5em}
    \setlength{\limitedlinewidth}{\linewidth}
    \addtolength{\limitedlinewidth}{-1em}
    \parshape 2
            0.em  \limitedheadlinewidth
            0.5em \limitedlinewidth
  }
  {}% indent
  {\bfseries}% header font
  {.}% punctuation
  {.5em}% after theorem header
  {\thmname{#1}\thmnumber{ #2}\thmnote{ (\textit{#3})}}% header specification (empty for default)

\theoremstyle{2maz_definition}
\newtheorem*{definition}{Definition}	
\newtheorem*{assumption}{Assumption}
\makeatother

\definecolor{ontogray}{RGB}{80,80,80}
\newcommand{\ontoKey}[1]{\textit{\textcolor{ontogray}{#1}}}

\usepackage{tabu}
\usepackage{booktabs}
\usepackage{colortbl}
\definecolor{lightgray}{RGB}{230,230,230}
\definecolor{midgray}{RGB}{170,170,170}
\definecolor{black}{RGB}{0,0,0}
\colorlet{headercolor}{lightgray}
\newenvironment{2maztabular}[1]
    {\begin{tabular}{#1}
        \toprule
        \rowcolor{lightgray}
    }
    {\bottomrule
    \end{tabular}}

\usepackage{xifthen}
\usepackage[final,xcolor=dvipdf]{changes}

\begin{document}
\title{Active Exploitation of Redundancies in Reconfigurable Multi-Robot Systems}
\author{Thomas M. Roehr$^{1}$
%~and~Frank Kirchner$^{1,2}$% <-this % stops a space
\thanks{$^{1}$DFKI GmbH Robotics Innovation Center Bremen, Robert-Hooke-Str. 1, 28359 Bremen, Germany, correspondence: thomas.roehr@dfki.de}% <-this % stops a space
%\thanks{$^{2}$University of Bremen, Germany}
}%


\markboth{}
{Roehr: Active Exploitation of Redundancies in
Reconfigurable Multi-Robot Systems}


\maketitle

\begin{abstract}
While traditional robotic systems come with a monolithic system design,
reconfigurable multi-robot systems can share and shift physical resources
in an on-demand fashion.
Multi-robot operations can benefit from this flexibility by actively managing system redundancies depending on current tasks and having more options to respond to failure events.
To support this active exploitation of redundancies in robotic systems, 
this paper details an organization model as basis for planning with reconfigurable
multi-robot systems. 
The model allows to exploit redundancies when optimizing a multi-robot system's probability of survival with respect to a desired mission.
The resulting planning approach trades safety against
efficiency in robotic operations and thereby offers a new perspective and tool to design and improve multi-robot missions.
We use a simulated multi-robot planetary exploration mission to evaluate this approach and highlight an exemplary performance landscape.
Our implementation of the organization model is open-source available (https://github.com/rock-knowledge-reasoning/knowledge-reasoning-moreorg).
\end{abstract}


% \fi
\IEEEpeerreviewmaketitle

\section{Introduction}
\IEEEPARstart{R}{econfigurable} multi-robot systems introduce a new dimension to
the design of future robot missions since they permit robots to exchange physical subsystems.
This flexibility to shift subsystems can be exploited to
actively manage the level of redundancy of individual robots.
This is especially interesting costly planetary space operations, which
require highly redundant robots.
The state of the art in planetary space missions are, however, single robotic systems.
Despite the fact that
international space agencies operate with multiple rovers on the same planet, cooperation between these system has not been targeted. With the consideration of building up
infrastructure, creating habitats to prepare human presence and supporting safe operations, 
this paradigm will have to shift.

\begin{figure}[b]
    \centering
    \includegraphics[width=\columnwidth]{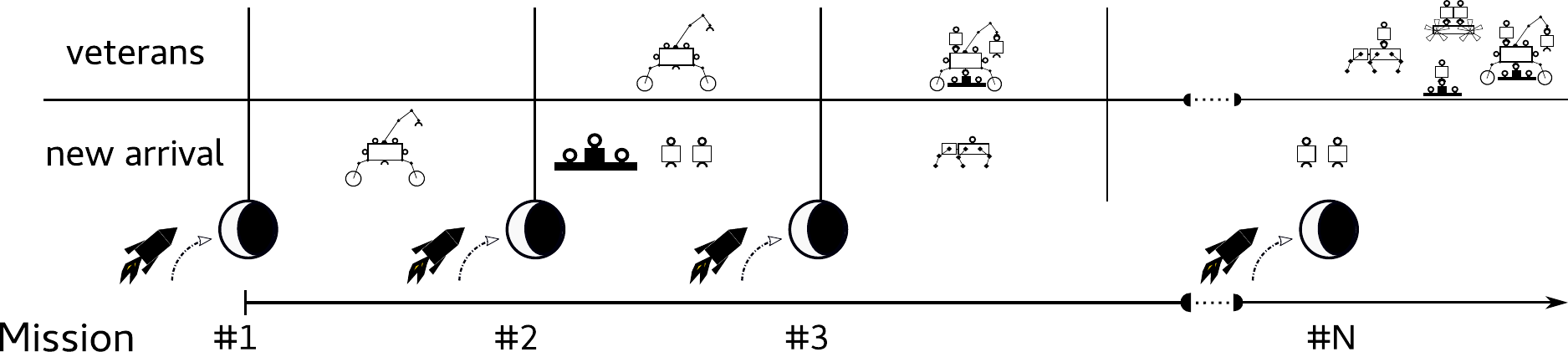}
    \caption{Schematic description of an incremental design of planetary space
    missions using reconfigurable multi-robot systems.}
    \label{fig:reconfigurable_mrs:incremental_mission_design}
\end{figure}

Current planetary space operations have to rely on ground operators for adaptations and repair,
which leads to a very slow and costly process.
The dependency on earth-based maintenance, or even hardware deliveries
should be minimized for future long-term space mission to achieve a 
An incremental mission design offers an alternative and the concept is depicted in Figure~\ref{fig:reconfigurable_mrs:incremental_mission_design}.
Here, not only software, but also hardware subsystems can evolve with the experience made in previous missions and incrementally improve already operating multi-agent systems.
The possibility to extend or refurbish existing hardware is a significant
advantage, even more when a team of robots can perform this process autonomously.
The autonomous exploitation of features of a reconfigurable multi-robot system is, however, a significant challenge including practical issues regarding distributed communication and planning approaches. Additionally, an application in a space context requires risk mitigation strategies and high safety standards.
Therefore, enabling planning approaches that permit to exploit redundancy and sharing of resources between robotic systems will be one step forward towards safe long-term operations of autonomous multi-robot systems.
By introducing a \gls{MoreOrg} in this paper, we offer a modelling approach with focus on (physically) reconfigurable multi-robot systems, although the model can embed classical non-reconfigurable robots.
This enables a planning approach which exploits reconfigurability as described in Section~\ref{sec:planning}. The focus of this paper is, however, on the design and role of the organization model in this context.

The organization modeling and planning approach for
    reconfigurable multi-robot systems finds its initial motivation in space
    applications. In this paper we also refer to this context to provide an
    exemplary use of our suggested modelling approach.  Meanwhile, the
    organization model builds on ontologies and is therefore extensible , so
    that users can add new subsystems, functionalities, robots, custom
    properties and inference rules by extending the ontological description.

\subsection{History and Related Work}
Implementations of reconfigurable multi-robot systems exist within a spectrum ranging from
industrial robots which allow an end effector exchange to fully self-reconfigurable
multi-robots systems \cite{Brandt:2007:ATRON,Kurokawa:2007:MTRAN,Davey:2012:SMORES}.
Research in the area of reconfigurable multi-robot systems has initially been
driven by the latter, i.e., the concept of so-called self-reconfigurable systems.
Their main design characteristic is a high-level of redundancy of mostly homogeneous modules, which can automatically restructure to establish a target structure; a broad review of self-reconfigurable multi-robot systems is provided by \textcite{Chennareddy:2017:SurveyRMRS} and \textcite{Liu:2016:SurveyRMRS}.
Self-reconfiguration aims a providing highly resilient systems, i.e., being able to recover from disruptive (structural) changes and failures.
These highly-modular systems typically suffer, however, from limited capabilities and thus lack broad applicability.
Planning approaches in this context focus on the transition between two organization states, e.g., \textcite{Baca:2014:ModRED} apply coalition game theory to optimize the organizational state. They characterize coalitions, however, based on strong assumptions with a utility function which
\begin{inparaenum}[(a)]
\item has a static utility for each agent independent of the coalition it will be embedded into, and
\item cannot account for interface compatibility issues leading to constrained coalition formation.
\end{inparaenum}
The swarm-bot system developed by \textcite{Mondada:2005:SBots} initially takes a middle ground and uses simple structured, yet reconfigurable swarm-based system in combination with a behavior-based control approach to exploit reconfigurability. They are able to illustrate team capabilities that arise from superaddition such as gap and obstacle traversal as well as (heavy) object transport.

Similarly to the swarm-bot system, our work targets reconfigurable multi-robot systems which consist of individual agents that can already be considered capable robots.
\textcite{Wilcox:2007:ATHLETE}, for instance, developed the reconfigurable six-legged robot ATHLETE to support infrastructure build up on planetary surfaces. Although \textcite{Wilcox:2007:ATHLETE} did approach automation of reconfiguration procedures, they did not develop high-level planning \added{approaches} to fully exploit reconfigurability. For similar capable robots reconfigurability focuses on the adaption of internal subsystems, e.g., the Scarab rover~\cite{Bartlett:2008:Scarab} is able to adapt its locomotion platform.
\textcite{Reid:2019:JFR} show how to exploit this kind of reconfigurability with a dedicated sampling-based (motion) planning approach after modeling the reconfiguration space of the motion planning system.

In the context of organization research outside of the robotics domain \textcite{Dignum:2009:Handbook} looks at reconfiguration of organizational processes involving humans.
According to Dignum the general need to actively organize teams aims at increasing
efficiency, and she sees flexible and adaptive organizations as suitable means to
deal with dynamic environments. 
She suggests that organizations conditionally adapt and should reorganize if this will lead to an increasing
success of an organization; even a suboptimal reorganization can be better than no
response at all.
However, the question when to reorganize and when to accept loss is left
unanswered.

In Dignum's work she points to \emph{strategic flexibility}, a concept
developed in the scope of managing high-technology industries by
\textcite{Evans:1991:StrategicFlexibility}. 
Evans framework conceptualizes the strategic use of a company's or
more generically a market player's flexibility.
Flexibility to adapt leads to a significant competitive advantage, since it
offers a market player additional means to encounter unforeseen events.
Hence, adaptation can directly lead to a
greater probability of survival or net monetary benefit for market players.
%According to Evans the use of a player's manoeuvres can be classified along two dimensions:
%temporal and intentional.
%Table~\ref{tab:reconfigurable_mrs:manoveures_characterisation} illustrates the
%resulting two dimensional matrix and the categorisation.
%\begin{table}[ht!]
%\caption{Characterisation of manoeuvres according to
%    \textcite{Evans:1991:StrategicFlexibility}.}
%\label{tab:reconfigurable_mrs:manoveures_characterisation}
%\centering
%\footnotesize
%\begin{tabular}{cccc}
%     &  & \multicolumn{2}{c}{\textit{temporal}} \\
%        &   & \textbf{proactive}            &\textbf{reactive} \\\cmidrule{3-4}\cmidrule{2-4}
%\multirow{2}{*}{\textit{intentional}}
%		& \textbf{offensive} & preemptive & exploitive \\
%		& \textbf{defensive} & protective  & corrective \\\cmidrule{2-4}
%\end{tabular}
%\end{table}
In his work \textcite{Evans:1991:StrategicFlexibility} refers to proactive, reactive, defensive and exploitative system capabilities and relates defensive one to robustness and resilience.
While robustness refers to a system which can endure impacts up to a
certain degree without breaking, resilience results from the ability to recover
from error and return into a functional state.

Especially resilience is a key to survival, and not only in natural systems,
but for technical and social systems alike shown by examples collected from \textcite{Zolli:2012:Resilience}.
Resilient systems, however, \added{rely on their capability to adapt}.
Therefore, reconfigurability can contribute to an increased resilience of
robotic systems.
Evans' conceptual framework is general enough to be applied to reconfigurable multi-robot systems, and his categorization of maneuvers
can be similarly applied for a characterization of robotic activities:
protective and corrective activities count as defensive maneuvers.

The design of a space robots is typically focusing on defensive measures by adding redundancies and \added{preparing} failure handling strategies.
What the flexibility of reconfigurable multi-robot systems offers, however, is the possibility for an active management of these redundancies, for instance, to adapt the organization to respond to functional requirements or to optimize the redundancy level across all available systems.
An active management with a global optimization policy will treat all resources equally. This means, that the controller of a robotic mission can influence the level of resources redundancies.
As a side effect \added{active management} might even result in an overall cost-optimized system design, by reducing the \added{mean} redundancy level \added{of the multi-robot system}.

Continuous optimization of an organization structure can also be found in \gls{MOISE+}. 
\citeauthor{MOISE+} focus on a design pattern to control the reconfiguration process and identify key components.
They outline an architecture to continuously optimize an organization structure
to achieve main organization's objectives. Objectives for a (sub)team can be
defined as \gls{HTN} in a so-called scheme, which leads to the definition of a behavioural pattern, e.g., they use playing soccer as primary example.
A reconfiguration process or rather transition from one team structure to another can be planned or unplanned: planned transitions can be triggered in a top-down fashion by an external operator, or they can be scheduled for a specific time. 
\citeauthor{Huebner:2004:MOISE} require planned transitions to follow a
previously defined and therefore static reorganization pattern, while unplanned
transitions have to be dynamically controlled by agents.

The reorganization process in MOISE+ itself requires forming a special, predefined group
structure: one agent has to adopt the role of the so-called \emph{OrgManager} in
order to organize the overall reconfiguration. The reconfiguration group
also requires at least one agent to take over the \emph{Designer} role, in order
to analyse the current status of the organizational structure, and suggest a potentially better structure.

In the area of robotics \gls{OMACS} is another approach for designing an organization model presented by
\textcite{Deloach:2008:AdaptiveOrganization, DeLoach:2009:OMACS}.
The main concepts in \gls{OMACS} are goals, roles, agents and
capabilities. \gls{OMACS} uses a capability-based representation for a role,
i.e., a role is defined by a set of capabilities. The quality of an
agent's capability can be quantified using normalized values.
In the same way \citeauthor{DeLoach:2009:OMACS} quantifies an agent's ability to fulfill a role based on its capabilities.
\gls{OMACS} sets the main focus on the quantification of the potential of abstract roles and agents to contribute towards an
organization's success. The usage of this information allows to optimize the team structure by allowing the best suited agent to handle a task, and thereby increases the likelihood of an organization's success.

Similar to MOISE+, \citeauthor{DeLoach:2009:OMACS} suggests the use of behaviour policies 
to control the cooperative behaviour of agents.
In \gls{OMACS} an organization designer can explicitly define reorganization
rules. For instance to specify if and how one agent can replace another agent
once the latter becomes unable to fulfil a role.
An application of runtime reorganization has been shown with three real robots, and a
single laptop agent by \textcite{Zhong:2011:OMACS}.

\gls{OMACS} assumes atomic capabilities without composition, and the value
normalisation to $[0,1]$ restricts the quantification, e.g., for a qualification
of capability, to a single dimension. The quality of an agent's capability has therefore (initially) unclear semantics, which limits the applicability of the approach in practical applications.

Our approach looks superficially very similar to \gls{MOISE+} and \gls{OMACS}
with respect to exploiting a mapping between structure and function. However,
\gls{MOISE+} as well as \gls{OMACS} do this still on a higher level:
\gls{MOISE+} simply defines the suitability of an agent to fill a role, while
\gls{OMACS} already analyses agent capabilities. Both organization models fit
loosely-coupled agent teams, but do neither take physical reconfiguration under
resource constraints into account, nor can they infer functionality of newly
composed agents. 
While \gls{MoreOrg} takes a similar capability-based approach of identifying an agent's available functionality to \gls{OMACS}, it
\begin{inparaenum}[(a)] 
\item derives this information dynamically from the available set of hardware and software resources, and thus permits inference of properties, 
\item does not (yet) characterize the quality of a function.
\end{inparaenum}
Instead we estimate the probability of survival of a function based on the available resources.
Neither \gls{MOISE+} nor \gls{OMACS} offer multi-agent planning, instead they
allow to control agent behaviour via predefined tasks.

The robotic framework KnowRob~\cite{Tenorth:2013:KnowRob} uses a semantic modelling approach and uses ontologies to represent the structure of a robotic system to create a mapping between structure and function. \textcite{Beetz:2010:CRAM} exploit this abstraction by defining plan templates for a single robot, which can then be used to identify required functionality during plan execution.

In contrast to the existing robot modeling and planning approaches, \gls{MoreOrg} can model a heterogeneous set of physically reconfigurable robots and infer agent as well as organization properties.
This permits an exploitation of superadditive effects and in parallel accounts for safety.
To the best of our knowledge none of the existing organization modeling and planning approaches in robotics has covered these aspects so far.

\subsection{Relation to Previous Work \& Contribution}
We base the results in this paper on the practical experience gained from working with multiple teams of reconfigurable systems \cite{Cordes:2011:LUNARES, Bartsch:2010:LUNARES, Roehr:2014:RIMRES, Sonsalla:2017:FTUTAH}\footnote{Field Trials Utah: \url{https://www.youtube.com/watch?v=pvKIzldni68}}.
Furthermore, this work is part of the planning approach developed with a special focus on reconfigurable multi-robot systems \cite{Roehr:2016:SpatioTemporalPlanning,Roehr:2018:IBERAMIA}.
The contributions of this paper are:
\begin{inparaenum}[(i)]
\item detailing an organization model for reconfigurable multi-robot systems with focus on functionality-based probability of survival, 
\item offering the open-source implementation of this model,
and
\item evaluating the trade-off between safety, efficacy and efficiency for an exemplary space mission.
\end{inparaenum}

\subsection{Outline}
This paper takes a bottom-up approach in describing \gls{MoreOrg}.
In Section~\ref{sec:rmrs} we first provide our revised formalization and terminology for reconfigurable multi-robot systems including atomic and composite agents.
Section~\ref{sec:organisation_modelling} extends the modeling and formalization
to agent and organization properties and in particular how they can be
generically inferred. Redundancy is a special properties and quantified in this context with respect to required functionality.
In Section~\ref{sec:planning} we provide details of our planning or rather optimization approach for reconfigurable multi-robot systems that is based on \gls{MoreOrg}.
In Section~\ref{sec:evaluation} we describe an exemplary planning result and in 
Section~\ref{sec:discussion} we discuss the current state, open challenges and give a critical review on our take on dealing with reconfigurable multi-robot systems.

\section{Reconfigurable Multi-Robot Systems}\label{sec:rmrs}
This section provides the basic notation, definitions and the underlying assumptions
regarding reconfigurable multi-robot systems.
The notation builds on the formalisms found in coalition
games~\cite{Weiss:2013:MultiagentSystems}.
In particular, the agent-type representation is based on the representations developed
by Shrot et al.~\cite{Shrot:2010:CoalitionFormationProblems} and Ueda et al.~\cite{Ueda:2011:CCF}.

While reconfiguration affects
hardware and software alike, the focus of this work is on handling physical reconfiguration of agents.
The level of granularity is chosen correspondingly
% Atomic agent
with the lowest level being a
physical agent which cannot be separated further into two or more physical agents.
This agent is denoted by \textbf{atomic agent}.
% Atomic agent
\begin{definition}[Atomic agent]\label{definition:atomic_agent}
An \textbf{atomic agent} $a$ represents a monolithic physical robotic system.
\end{definition}
Note that a physical agent representing an atomic agent still contains
subsystems.
They are, however, inseparable parts of the physical agent.
\begin{definition}[(Atomic) Agent pool]\label{definition:agent_pool}
An \textbf{agent pool} $A$ denotes a set of atomic atomic agents, such that
$A=\{a_1,\ldots,a_{|A|}\}$, is the set of all atomic agents, and $a \in A$ or
equivalently $\{ a \} \subseteq A$.
A set of agents pools is denoted by $\mathbf{A} = \{A_1,\dots,A_{|\mathbf{A}|}\}$.
\end{definition}

Connection interfaces are the key elements in reconfigurable system and here, open the
opportunity for combining two or more atomic agents.
A composition from two or more atomic agents is referred to as
\textbf{composite agent}.
The join operator $\cup$ in the following definition for composite agents aligns well with the actual physical join operation of atomic agents and 
permits an intuitive representation.
% Composite Agent
\begin{definition}[Composite agent]\label{definition:composite_agent}
A linked system of two or more atomic agents is a set $CA$, which is denoted by
\textbf{composite agent} $CA = a_i \cup \ldots \cup a_j = \{a_i,\ldots,a_j\}$, where $a_i,\ldots,a_j \in A,
|A| \geq |CA| > 1$.
\end{definition}

\begin{figure}[h!t]
\centering
\includegraphics[width=\columnwidth]{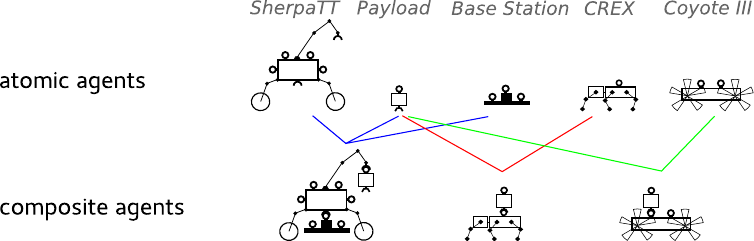}
\caption{An available set of atomic agents and a subset of composite agents that
can be formed by combining different atomic agents (see \cite{Roehr:2014:RIMRES,Sonsalla:2017:FTUTAH} for details on the real counterparts). From top left to right: (1) a rover with four male and two female interfaces, (2) a payload item with one male and one female interface, (3) a base station with four male interfaces, (3) a legged robot with one male interface and (4) a star-wheeled robot with two male interfaces }
\label{fig:reconfigurable_mrs:agent_pool}
\end{figure}

Figure~\ref{fig:reconfigurable_mrs:agent_pool} illustrates the general approach to agent
composition as basis for superaddition, as explained with the following example:
A mobile robot (atomic agent $m$) can share its power source with other robots,
but it has no camera. After attaching an unpowered atomic agent $c$ which has one camera
as a subsystem, the composite agent $\{m,c\}$ is equipped to take
images. It can now move to any location and
take images - a functionality neither of the atomic agents $m$ or $c$ provides.

Combinatorial explosion is one of the main challenges to deal with when
considering a reconfigurable system with a large number of atomic agents.
One means to reduce the effects of combinatorial explosion is typing.
Agent typing allows dealing with same typed agents using
homogeneously formed partitions of an overall set of agents.
\begin{definition}[Atomic and composite agent type]\label{definition:atomic_composite_agent_type}
The type of an atomic agent $a$ is denoted by $\hat{a}$ and equivalently for a
composite agent $CA$ the type is denoted by $\widehat{CA}$. The set of all atomic agent
types is denoted by $\theta(A) = \{ 1, \ldots, |\theta(A)| \}$, with
the corresponding type-partitioned sets of agent instances 
$A^{1},\ldots,A^{|\theta(A)|}$, where $A = A^{1} \cup \ldots \cup
A^{|\theta(A)|}$ and $A^{x}$ represents an agent pool containing only atomic agents of type $x$.
\end{definition}
The concept of a (general) agent wraps the concepts of atomic and composite agents.
Henceforth, the term agent is equivalently used to the term general agent.
%
% General agent
%\begin{definition}[General agent]\label{definition:general_agent}
%Any subset $A' \subseteq A$, where $A' \neq \emptyset$ forms a physical
%coalition through a set of links $L$ is denoted by \textbf{general agent}.
%A (general) agent $GA = (A,L)$ has a corresponding atomic agent type
%partitioned set of agent instances $A^{1}, \ldots, A^{|\theta(A)|}$, where $A =
%A^{1} \cup \ldots \cup A^{|\theta(A)|}$.
%$L = \{(a_{i,m},a_{j,n}),\dots\}$ represents the set of links between the atomic
%agents composing $GA$.
%\end{definition}
\begin{definition}[General agent]\label{definition:general_agent}
Any subset $A' \subseteq A$, where $A' \neq \emptyset$ forms a physical
coalition is denoted by \textbf{general agent}.
A (general) agent has a corresponding atomic agent type
partitioned set of agent instances $A^{1}, \ldots, A^{|\theta(A)|}$, where $A =
A^{1} \cup \ldots \cup A^{|\theta(A)|}$.
\end{definition}
%
%\begin{definition}
%A (general) agent type $\widehat{GA}$ is represented as a tuple set of agent type and type cardinality: $\widehat{GA} = \{(\hat{a}_k,c_k),\ldots,(\hat{a}_l,c_l)\}$, where $a_i \in A$ and $0 \leq c_i \leq |A^{\hat{a}_i}|$.
%$\widehat{GA} \supseteq \widehat{GA'} \iff \forall (\hat{a}_i,c_i)\in \widehat{GA},(\hat{a}_i,c_i')\in \widehat{GA'}: c_i \geq c_i', \text{ where } i = 1, \ldots ,|\widehat{A}|$. Such a tuple set will be denoted an \textbf{agent pool}.
%\end{definition}
%
% General agent type
%\glsadd{agent:type:general}
\begin{definition}[General agent type]\label{definition:general_agent_type}
The type of a (general) agent $GA$ is denoted by $\widehat{GA}$.
A general agent type $\widehat{GA}$ is represented as a function
$\gamma_{\widehat{GA}}: \theta(A) \rightarrow \mathbb{N}_0$.
The function $\gamma_{\widehat{GA}}$ maps an atomic agent type $\hat{a}$ to the
cardinality $c_{\hat{a}}$ of the type partition of $\widehat{GA}$, such that  $c_{\hat{a}} = |GA^{\hat{a}}|$.
Equivalently to $\gamma_{\widehat{GA}}(\hat{a}) \geq 1$ the following
notation will be used: $\hat{a} \in \widehat{GA}$, and $\hat{a} \not \in
\widehat{GA}$ for $\gamma_{\widehat{GA}}(\hat{a}) = 0$.

The reverse mapping from type $\widehat{GA}$ to the general agent is denoted by $i(\widehat{GA}) = GA$.
\end{definition}
A general agent type is also represented as a collection of tuples relating
agent type and cardinality: $\{
    (\hat{a}_0, c_{\hat{a}_0}),(\hat{a}_1,c_{\hat{a}_1}), \ldots,
    (\hat{a}_n,c_{\hat{a}_n})\}$.
\\
%
%\begin{definition}[Constructible agent types]\label{definition:constructible_agent_types}
%The set of all constructible general agent types from a set of atomic agents A is denoted 
%by $\Theta(A)$; it
%represents the collection of all general agent types that are found in the powerset of all agents $\mathcal{P}^A$.
%\end{definition}
%%\vspace{10px}
%
% equivalent they are also strategically equivalent''

An agent pool can now be represented in two ways:
\begin{inparaenum}[(i)]
\item as set of atomic agents as already introduced, or
\item as general agent type $\widehat{A}$, such that $\forall a \in A: \gamma_{\widehat{A}}(\hat{a}) = |A^{\hat{a}}|$,
\end{inparaenum}
where the latter offers a more compact representation and is used preferably in our implementation of the organization model.

To execute robotic missions, atomic agents from an available agent pool will be assigned to particular tasks.
%\glsadd{sym:agent:role}
However, if multiple atomic agents of the same type exist and equal start
conditions hold for these atomic agents, multiple equivalent assignments of atomic agents
to a task are possible.
For that purpose, requirements for atomic agents will be defined by so-called
roles, which act as correctly typed placeholders for instances of
an agent type.
\begin{definition}[Atomic agent role]\label{definition:agent_role}
An \textbf{atomic agent role} $r^{\hat{a}}$ 
represents an anonymous agent instance of an atomic agent type $\hat{a}$.
%A set of agent roles with a one-to-one mapping to an agent pool $A$ is denoted
%by $r(A)$.
\end{definition}
Given an overall set of atomic agents, various reconfiguration states of the
overall systems are possible. These reconfiguration states result from forming
different sets of composite agents, but always with the restriction of the
overall available set of atomic agents.
%\glsadd{sym:coalition:structure}
In the field of multi-agent systems and particularly characteristic
function games (see \cite{Weiss:2013:MultiagentSystems}) this leads to so-called coalition
structures. A coalition structure represents the set of active atomic and composite agents that form
a reconfigurable multi-robot system. Note that we will also use the term \emph{organization} in order to describe a reconfigurable multi-robot system represented by an agent pool $A$.
%\begin{definition}
%A resource $r \in R $ represents a resource, where $R = {r_0,\ldots,r_n}$ is the set of all resources
%\end{definition}
\begin{definition}[Coalition structure]\label{definition:coalition_structure}
A coalition structure of an agent pool A is denoted by $CS^A$ and is represented by
a set of disjunct general agents $CS^A = \{ GA_0, \ldots,
GA_n \}$, where $GA_0 \cup \ldots \cup GA_n = A$, and $i,j = 0, \ldots,
n, \forall i,j,i \neq j: GA_i \cap GA_j = \emptyset$.
\end{definition}
%\begin{definition}
%A coalition type structure of an agent set A is denoted $\widehat{CS_A}$ and represents a set of general agent types, i.e., 
%\end{definition}
%
Composite agents result from the combination of atomic agents.
We use the following definitions to separate the current
(realized and physically assembled) set of general agents in a coalition
structure from the (virtual) set of agents, which can be formed from the set of
atomic agents.
%\glsadd{operative agent}
%\glsadd{dormant agent}
\begin{definition}[Operative and dormant agents]\label{definition:operative_agents}
Let the current state of a reconfigurable multi-robot system be described by
a coalition structure $CS^{A}$.
Then all general agents $GA \in CS^{A}$ are referred to as \textbf{operative agents},
and complementary, all general agents $GA \in \mathcal{P}^{A} \land GA \not \in
CS$ are referred to as \textbf{dormant agents}, where $\mathcal{P}^{A}$ is the powerset of all atomic agents.
\end{definition}
The previous definitions look at a reconfigurable multi-robot system as a collection of agents, 
and consider pairing and coalitions only at this level of modularity.
A reconfigurable multi-robot system can form composite agents in
different ways depending upon the compatibility of interfaces.
Hence, to perform a detailed reasoning on connectivity of agents, we also have to account for the physical interfaces as subsystems of an agent to analyze the feasibility of all agents.
composite agent.
The scope of the presented formal description so far, is based on a set-theory description and covers what is denoted agent space.
\begin{definition}[Agent space]\label{definition:agent_space}
\textbf{Agent space} denotes the set-theory based view to a reconfigurable multi-robot
system without constraining the connectivity between any two agents.
\end{definition}

Agent space is, however, only a restricted view onto link space.
\begin{definition}[Link space]\label{definition:link_space}
\textbf{Link space} denotes a graph-based structure of a reconfigurable multi-robot system.
In link space a reconfigurable multi-robot system is represented by an
undirected graph $G=(V,E)$, where each vertex $v \in V$ maps to an atomic agent's interface and an edge $e =
 (u,v)$, $u,v \in V$ represents the existing connection between two interfaces.
\end{definition}

The current modeling approach regarding link space accounts only for edges that represent electromechanical connections between agents as precondition for sharing resource in a composite agent. Data connections between software and hardware components as well as configuration options of hardware and software components are currently left out in our modeling approach.

\subsection{Assumptions}
A large spectrum of reconfigurable
multi-robot systems exists.
Most often, fully distributed control approaches
apply, due to the use of swarm-based systems.
The definition of the general agent already reflects one important design
choice of this work, which relaxes this apparent
requirement for distribution.
Instead of enforcing distributed control approaches at all system levels,
centralized control approaches for locally autonomous and self-sustained operation of agents
are permitted and feasible. 
This implicitly allows an atomic agent to act as
a temporary 'master' in a master-slave architecture.
When forming a composite agent, for instance, a single atomic agent in this
formation acts as master, which is able to control all other attached atomic
agents.
In effect, each general agent represents a single-minded (collaborative)
agent.
The distribution of the overall agent system is still maintained by an
appropriate design of the operational infrastructure.
%While many approaches towards self-reconfigurable multi-robot systems apply
%distributed control approaches, the implementation of the reference systems aims
%at the dynamic creation of composite, but tightly coupled agents, which can rely
%on centralised control approaches.
\begin{assumption}[Individual agent]
Each atomic and composite agent comprises a central controller and thus
represents an individual, single-minded agent.
\end{assumption}
Generally, two atomic agents can connect via multiple interfaces.
We assume, however, limited connectivity
and currently do not consider geometrical constraints.
This restriction allows us to focus on the identification
of essential needs for modeling and automating of reconfigurable multi-robot systems.
\begin{assumption}[Single link]\label{assumption:mechanical_coupling}
A \added{physical} coupling between two atomic agents can only be established through two and only two compatible coupling interfaces.
\end{assumption}

In principle Definition~\ref{definition:atomic_composite_agent_type} allows a
single agent to have multiple types.
However, we assume a single characterising agent type.
Meanwhile, one agent type can still inherit the properties of a parent type.
\begin{assumption}[Single agent type]\label{assumption:single_agent_type}
An agent can be mapped to a single agent type only.
\end{assumption}
%
%\glsadd{agent:type:inheritance}
\begin{assumption}[Agent type inheritance]\label{assumption:agent_type_inheritance}
An agent type can inherit the properties of another agent type.
\end{assumption}

A key assumption, when dealing with an active exploitation of resource is the possibility to join the available resources of two or more agents.
Hence, when two or more atomic agents form a composite agent, they join their set of
resources.
In principle, geometrical restrictions might apply to reuse the set of resources effectively. 
However, we initially assume that resources are shared without restriction within a composite agent.
\begin{assumption}[Resource usage]\label{assumption:resource_usage}
A composite agent can reuse the subsystems of its composing atomic agents.
\end{assumption}

To enable resource sharing in a composite agent, various ways of coupling two or more atomic agents can be considered, e.g., electromechanical or thermo-electromechanical.
We currently assume, however, that composite agents establish links between their composing atomic agents which permit data, energy and power transmission.
\begin{assumption}[Agent Linkage]\label{assumption:agent_linkage}
Links which connect atomic agents in a composite agent permit transfer of data, energy and power.
\end{assumption}

To effectively exploit resources in redundant structure, the following assumption is made:
\begin{assumption}[Component substitution]\label{assumption:substitution}
To maintain the functionality of an agent, one component can replace another if it
is an instance of the other's class, which also includes instances of subclasses.
\end{assumption}
This seems like a strong assumption, since even if components
are instances of the same concept, e.g., a camera, it might not be possible to
substitute one with the other without loosing functionality.
However, this is a matter of modeling equivalence as part of the ontological design in \gls{MoreOrg}.
%considers a shared use of resources in a composite agent.

\section{Organization Modelling}\label{sec:organisation_modelling}
\begin{figure}
\centering
\includegraphics[width=\columnwidth]{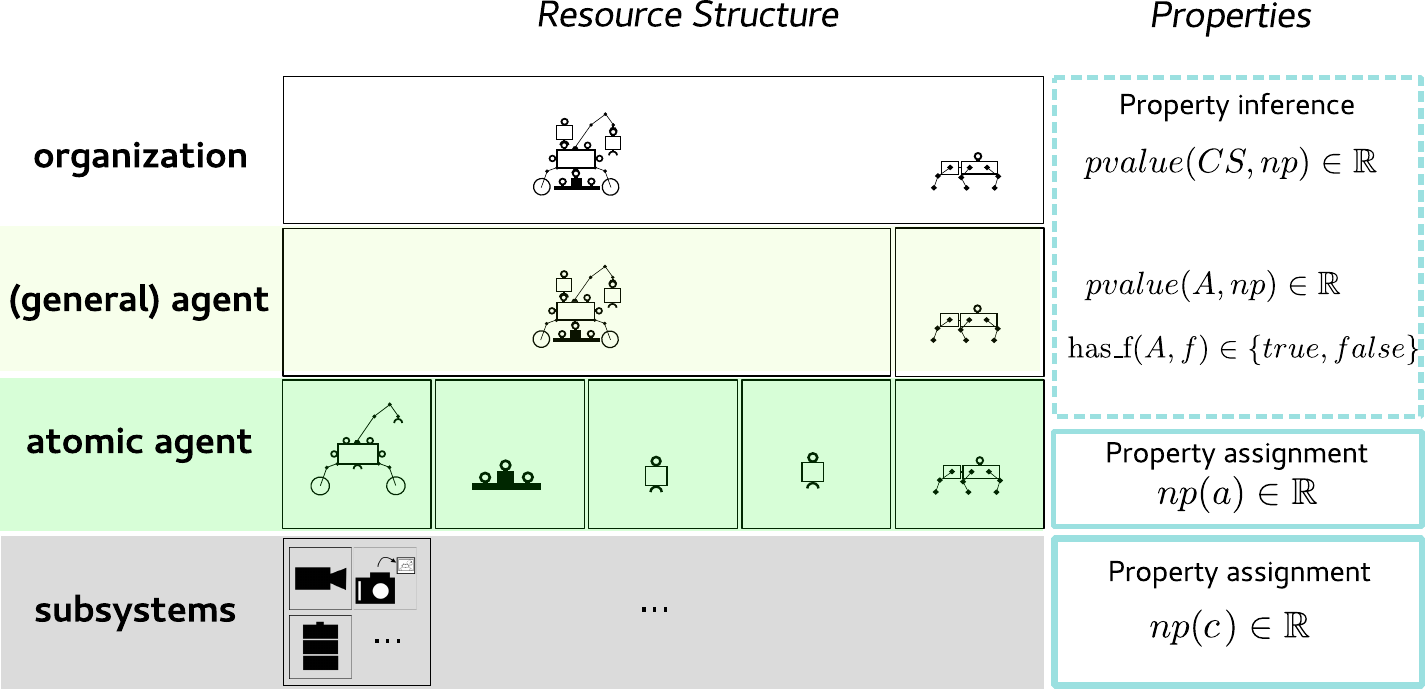}
\caption{The organization model is based on a hierarchical view and corresponding property generation. Atomic agents come with (mainly) static properties assignments, while composite agents and overall organization properties have to be dynamically derived from the active coalition structure.}
\label{fig:organization_modelling:property_levels}
\end{figure}

We have developed the organization model \gls{MoreOrg}\footnote{\url{http://github.com/rock-knowledge-reasoning/knowledge-reasoning-moreorg}} to quantify the properties of a reconfigurable multi-robot system and provide cost measures for a reconfigurable multi-robot system.
The organization model permits a quantification of system properties at different granularity levels and can characterize the active set of agents, i.e., the coalition structure of the organization by using a bottom-up approach.
Figure~\ref{fig:organization_modelling:property_levels} depicts the hierarchical decomposition of an organization which serves as baseline for \gls{MoreOrg}'s reasoning approach.
The coalition structure of a reconfigurable multi-robot system can change on-demand, which might involve creating and/or removing one or more connection between atomic agents, but all agents can be characterized by their set of associated resources, i.e., hardware and software components.

\gls{MoreOrg} relies on ontologies to describe resources in general
and more specifically atomic agents and their associated
functionalities and subsystems. All subsystems and atomic agents are
characterized by data properties, e.g., \gls{MoreOrg} focuses on a set of
numeric data properties to enable mobile transport agents (cf.
\cite{Roehr:2018:IBERAMIA} where we relate planning for reconfigurable
multi-robot systems to Vehicle Routing Problems).
As a benefit of this ontological representation, \gls{DL}-based reasoning can be applied and an agent's available functionalities and properties can be inferred from its composing set of resources or rather subsystems. 
Additional reasoning mechanisms are applied to finally describe the organizational properties, where our focus is set on efficacy and safety.
The following subsections detail the organization model and its reasoning approach.

\subsection{Subsystem Properties}
Subsystems are tightly bound to atomic agents and come with statically defined properties. All subsystems are at least associated with a \emph{probability of survival} to define their reliability.
This is the basis for computing a safety measure for the overall multi-robot organization.
The details of this computation are provided in Section~\ref{sec:organisation_properties}.

\subsection{Generic Agent Type Properties}\label{chapter:organisation_modelling:composite_agent:properties}
\gls{MoreOrg} provides a mechanism to defined agent properties in a general way,
such that composite agent properties can be derived from the atomic agents that
form the agent.

\begin{definition}[Agent Type Property Value]
The value function for an agent type $\widehat{A}$ and a numeric property named $np$ is denoted by $pvalue(\widehat{A},np)$.
\end{definition}
Some atomic agent type properties as listed in
Table~\ref{tab:organisation_model:agent_type_static_properties} are statically
defined and required to implement the planning approach. Note that other uses of
the organization model can lead to the use of a completely different set of
properties.
Hence, the listed properties are examples only and have been added to support the operation of a logistics chain as targeted reconfigurable multi-robot planning problem in space exploration (see Section~\ref{sec:planning}).
Nevertheless, the already defined properties will likely be sufficient and
needed for many standard robotic scenarios.
Any missing properties can easily be added to the ontology if needed.

While many properties of atomic agent types are directly set, 
some atomic agents' properties and all properties of composite agents have to be inferred.
Here, \gls{MoreOrg} distinguishes between boolean and numeric properties (cf. Table~\ref{tab:organisation_model:agent_type_inferred_properties}).

\begin{table*}
\centering
\footnotesize
\caption{Static atomic agent type properties (adapted from \cite{Roehr:2018:IBERAMIA})}
\label{tab:organisation_model:agent_type_static_properties}
\begin{2maztabular}{llp{8.5cm}}
\textbf{Property} & \textbf{Syntax} & \textbf{Description} \\\midrule

\textbf{velocity} & $v_{nom}(\widehat{a})$ & nominal velocity of an agent type
$\widehat{a}$, $|v_{nom}| > 0$ for mobile atomic agent types and $v_{nom} = 0$ for immobile \\

\textbf{transport capacity} & $tcap(\hat{a})$ & maximum total
    capacity of an agent of type $\hat{a}$ to transport other agents
    %$acap(\widehat{a_i}, \widehat{a_j})$ defines the maximum capacity of an agent
    %type $\widehat{a_i}$ to transport an agent type $\widehat{a_j}$, it defaults
    %to the maximum total capacity
    \\
\textbf{transport consumption} & $tcon(\hat{a})$ & number
    of storage units an agent of type $\hat{a}$ consumes, when being
    transported by another agent ($tcon$ is set to 1 for all agent
    types if not mentioned otherwise) \\
\textbf{transport load} & $tload(a)$ & current load transported
    by an atomic agent $a$, i.e., represents the consumed transport capacity of
    an agent \\
\textbf{power source capacity} & $esourcecap(\hat{a})$ & power source capacity of an atomic agent in Ah \\
\textbf{supply voltage} & $esupply(\hat{a})$ & electrical supply voltage of an atomic agent in V \\
\textbf{power consumption} & $pw(\hat{a})$ & (electrical) power consumption of an
    agent of type $\hat{a}$ \\
%\item[\textbf{safety metric}] $saf(s), s \in STR$,defines the safety metric
%    associated with an \gls{stqe} based on the available general agent and with
%    respect to the required set of functions; currently a redundancy based
%    model is used to estimate the probability of survival based on an agent's
%    set of component required to provide the functionalities in $\mathcal{F}$
%    (cf. \parencite{Roehr:2016:SpatioTemporalPlanning} for details), such that $0 \leq saf(s) \leq 1$.
\end{2maztabular}
\end{table*}
\begin{table*}
\centering
\footnotesize
\caption{Inferred agent type properties}
\label{tab:organisation_model:agent_type_inferred_properties}
\begin{2maztabular}{llp{8.5cm}}
\textbf{Property} & \textbf{Syntax} & \textbf{Description} \\\midrule
\textbf{has functionality} & $has\_f(\widehat{A},f)$ & boolean property, defines whether an
    agent of type $\hat{A}$ has functionality $f$. The truth value is inferred from the resource dependencies that are defined for $f$ in the ontology. \\
\textbf{efficacy} &  $\textit{efficacy}(\widehat{A}, \mathcal{F})$ &  boolean property, defines whether an agent type $\widehat{A}$ supports all functionalities in $\mathcal{F}$ or not \\\midrule
\textbf{numeric property} & $pvalue(\widehat{A},np)$ & numeric property, value for an agent of type $\widehat{A}$ and a property $np$ \\
\textbf{reliability} & $R(\widehat{A}, \mathcal{F})$ & numeric property, reliability ([0,1]) of an agent type $\widehat{A}$ with respect to functionalities in $\mathcal{F}$, based on resource redundancies \\
\textbf{operation cost} & \textit{ocost}$(\widehat{A},t)$ & numeric property, here: total consumed power for an agent type $\widehat{A}$ over time $t$ \\
%
%        
%\textbf{autonomous mobile} & $amobile(\hat{a})$ & boolean property which defines whether an
%    agent of type $\hat{a}$ is mobile and can autonomously navigate or not
%    ($\lnot amobile(\hat{a})$). This predicate is inferred by \gls{MoreOrg} based on
%    the availability of capabilites \ontoKey{Locomotion}, \ontoKey{Mapping}, \ontoKey{MotionPlanning}, \ontoKey{SelfLocalization} and a subsystem \ontoKey{PowerSource}.
%    \\
%\textbf{power consumption} & $pw(\hat{a})$ & (electrical) power consumption of an
%    agent of type $\hat{a}$ \\
%\item[\textbf{safety metric}] $saf(s), s \in STR$,defines the safety metric
%    associated with an \gls{stqe} based on the available general agent and with
%    respect to the required set of functions; currently a redundancy based
%    model is used to estimate the probability of survival based on an agent's
%    set of component required to provide the functionalities in $\mathcal{F}$
%    (cf. \parencite{Roehr:2016:SpatioTemporalPlanning} for details), such that $0 \leq saf(s) \leq 1$.
\end{2maztabular}
\end{table*}

\subsubsection{Boolean Properties}
Boolean properties map to the availability of particular capabilities and they
can be inferred from available combinations of functionalities and subsystems,
e.g., the functionality \ontoKey{AutonomousNavigation} is inferred from the
availability of other capabilities and subsystems, here 
\ontoKey{Locomotion}, \ontoKey{Mapping}, \ontoKey{MotionPlanning}, \ontoKey{SelfLocalization} and a subsystem \ontoKey{PowerSource}.
The boolean property $has\_f(\widehat{A},f)$ defines whether an agent $A$ supports a functionality $f$ or not. 
This boolean property enables selection mechanisms, e.g., to identify agents which have the functionalities to perform a particular tasks.

For atomic agents the inference of available functionality is based on ontological reasoning and exploits available \gls{DL}-based reasoners, here Fact++ \cite{Tsarkov:2006:FACT}.
For composite agents, we allow to quantify the \textit{support} for a functionality by an agent's resources, so that the boolean property depends on the amount of support:
\begin{eqnarray}\label{equation:has_functionality}
\textit{has\_f}(\widehat{A},f) &=&
\begin{cases}
 \text{true} & \text{if }support(\hat{A},f) \geq 1 \\
 \text{false} & \text{otherwise} \\
\end{cases}
\end{eqnarray},
where $\widehat{A}$ is the agent type, $f$ a functionality and $support$ is defined in the following.

Support for an agent's functionality is based on a single resource concept $c$, e.g., where $c$ can represent a subsystem type such as \ontoKey{Camera}, as follows
(see also \parencite{Roehr:2016:SpatioTemporalPlanning}):
\begin{align}\label{equation:support_resource_atomic_agent}
    support(\widehat{A},c,f) &= \begin{cases}
                    0 & \text{if } card_{min}(c,f) = 0\\
                    \frac{card_{max}(c,\widehat{A})}{card_{min}(c,f)} & \text{ otherwise}
                \end{cases}\quad \text{,}
\end{align}
where $card_{min}$ and $card_{max}$ return the minimum required and maximum available cardinality of resource instances (including instances of derived resource concepts), respectively.
Accordingly, support of a functionality $f$ with respect to a resource class $c$ can
be categorized as follows:
% Consider using a definition here as well
\begin{eqnarray}\label{equation:support_categories}
support(\widehat{A},c, f) &=&
    \begin{cases}
        0 &  \text{no support}\\
        \geq 1 & \text{full support} \\
        > 0 \text{ and} < 1 & \text{partial support} \\
    \end{cases}
\end{eqnarray}
Support for a single functionality and subsequently for a set of functionalities $\mathcal{F}$ is then defined as:
\begin{align}
    support(\widehat{A},f) &= \operatorname*{min}_{c \in \mathcal{C}}
    support(\widehat{A}, c, f) \quad,
\end{align}
where $\mathcal{C}$ is a set of resource concepts and $\forall c \in \mathcal{C}:
card_{min}(c,f) \geq 1$ to account only for relevant resource concepts, and
\begin{align}
support(\widehat{A}, \mathcal{F}) = \min_{f \in \mathcal{F}} support(\widehat{A}, f)
\end{align}

\subsubsection{Numeric Properties}
Not all static numeric properties of atomic agents need to be directly assigned,
since they can be inferred from other properties, e.g., due to laws of physics.
As an example available energy capacity (Wh) can be derived from the capacity of the power source (Ah) and the supply voltage (V):

$pvalue(A,ecap) = $

\begin{tabular}{p{1cm}l}
 & $pvalue(A, esourcecap) \cdot pvalue(A,esupply)$ \text{, }
\end{tabular}
where $|A| = 1$.
\gls{MoreOrg} allows to define these mathematical relationships between
properties. This is done by annotating properties in the
ontology and by using a simple \gls{DSL} in combination with a math parser library~\footnote{muParser: \url{https://beltoforion.de/en/muparser}}.

To infer the value of numeric properties for composite agents, \gls{MoreOrg} permits the definition of custom inference rules.
These rules are defined as higher-order functions, which can be constructed from selection policies and composition operations.
\paragraph{Selection Policy}
A selection policy $A' = sel(A)$ permits to identify a subselection of atomic agents $A'$ from a (general) agent $A$ according to defined criteria. It is built from subselection policies which take the form: $\mathbf{A}' = sel(\mathbf{A}, \dots)$.
\gls{MoreOrg} offers three basic subselection policies, which to build custom selection policies:
\begin{compactenum}[1.]
\item agent size-based selection:
$\mathbf{A}' = \textit{size\_sel}(\mathbf{A},op,\beta)$, where $\forall A \in \mathbf{A}': |A|$ $op$ $\beta$ with $op \in \{<,<=,>,>=,=\}$
\item functionality-based selection:
$\mathbf{A}' = \textit{func\_sel}(\mathbf{A},f)$, where $\forall A \in \mathbf{A}': \textit{has\_f}(\widehat{A},f)$
\item property-based selection:
$\mathbf{A}' = \textit{prop\_sel}(\mathbf{A},op,np)$, where $\forall A \in
\mathbf{A}': A = op_{A \in \mathbf{A}}$ \textit{pvalue}$(\widehat{A},np)$, where $op \in \{ 
\operatorname*{argmax},  \operatorname*{argmin} \}$
\end{compactenum}
By chaining basic selection policies according to $f(\mathbf{A}) \circ g(\mathbf{A}) = f(g(\mathbf{A}))$, custom selection policies can be defined in the ontology.
The following example illustrates the policy to identify all transport providers with maximum transport capacity in a composite agent:

\textit{sel}$_{\ontoKey{TransportProvider}}(A) =$\\
\begin{tabular}{p{1cm}l}
 & \textit{random\_sel}$(\mathbf{A})$ \\
 &  $\circ$ \textit{prop\_sel}$(\mathbf{A},argmax,tcap)$\\
 &  $\circ$  \textit{size\_sel}$(\mathbf{A},=,1)$ \\
 &  $\circ$  \textit{func\_sel}$(\mathbf{A},$\ontoKey{TransportProvider}$)$ \\
 &  $\circ$ $\mathcal{P}^{A}$
\end{tabular}\\
, where $\mathcal{P}^{A}$ is the powerset of all atomic agents in A and
$random\_sel$ a tie-breaker function:

\begin{eqnarray}\label{equation:organisation_model:random_sel}
    \textit{random\_sel}(\mathbf{A}) &=& \begin{cases}
 \emptyset & \text{if } \mathbf{A} = \emptyset \\
 \text{randomly picked} & \text{otherwise} \\
 \text{ element from }\mathbf{A}
\end{cases}
\end{eqnarray}
The inverse selection policy is denoted by $\lnot sel(A) = A \setminus sel(A)$.\\

\paragraph{Composition operator}
A composition operator $c(A,np,op)$ combines the numeric property values of all atomic agents that form an agent.
Note that composition operators currently need to be hard-coded into the model.
The default supported operator is defined as:
\[
	c(A,np,+) = \sum_{a \in A} pvalue(\{ \hat{a} \},np)\text{ ,} 
\],
where $np$ is a numeric property.

\paragraph{Inference Rule}
Both, composition operators and selection policies can be combined to form an
inference rule for composite agents, e.g.,
for all properties relating to locomotion, such as nominal velocity:

$pvalue(\widehat{A},v_{nom}) =$

\begin{tabular}{p{1cm}l}
   & $c(sel_{\ontoKey{TransportProvider}}(i(\widehat{A})),v_{nom},+)$ \\
\end{tabular},
which maps to the $v_{nom}$ property value of the
only available \ontoKey{TransportProvider} or 0 if there is none.

More complex inference is required to compute the (remaining)
transport capacity:

	$pvalue(\widehat{A},tcap) =$

\begin{tabular}{p{1cm}l}
   & $c(sel_{\ontoKey{TransportProvider}}(i(\widehat{A})),tcap,+)$ \\
  & - $c(\lnot sel_{\ontoKey{TransportProvider}}(i(\widehat{A})),tcon,+)$ \\
\end{tabular}\\
, where $i(\widehat{A})$ represents the reverse mapping from type to agent.\\

Inference rules are defined in the ontology, so that users can add their own rules.

\subsection{Special Agent Type Properties}\label{sec:organisation_properties:safety}
Some special agent properties exist, where in contrast to the generic agent type properties the reasoning mechanisms are hard-coded into the model.

\subsubsection{Safety}
In \gls{MoreOrg} the computation of safety of an agent is a special property motivated through the space application context.
Safety is based on resource redundancy under the assumption of possible component substitution (see Section~\ref{assumption:substitution}).
The measure for redundancy is the central part of our safety heuristic and it is based
on the standard modeling approach for parallel and serial component-based
systems~(see \cite{Rausand:2009:ReliabilityTheory}).
Each resource can be associated with a probability of survival, so that an
overall probability of survival can be computed using a function decomposition
tree approach.
Information about the probability of survival of components should be derived from an
initial system identification and is ideally updated with 
performance and degradation information from the real system.

The reliability $R_f$, also referred to as probability of survival, of a single functionality $f$ can be
computed by accounting for parallel components, i.e., resources that are not
strictly required but which can serve as replacement:
\begin{align}
R_f(t) = \begin{cases}
1 - \prod_{i=1}^n (1-p_i(t)) &\text{parallel system} \\
\prod_{i=1}^n p_i(t) & \text{serial system} \\
\end{cases} \quad,
\end{align}
where $p_i(t)$ is the time-dependent probability of survival with $0 \leq p_i(t)
\leq 1$.
While component degrading can be one reason for a change of the probability of
survival, \gls{MoreOrg} leaves the use of time-dependence as future
improvement and instead uses a static probability of survival with $t=0$.

\begin{figure}
    \centering
    \includegraphics[height=2.5cm]{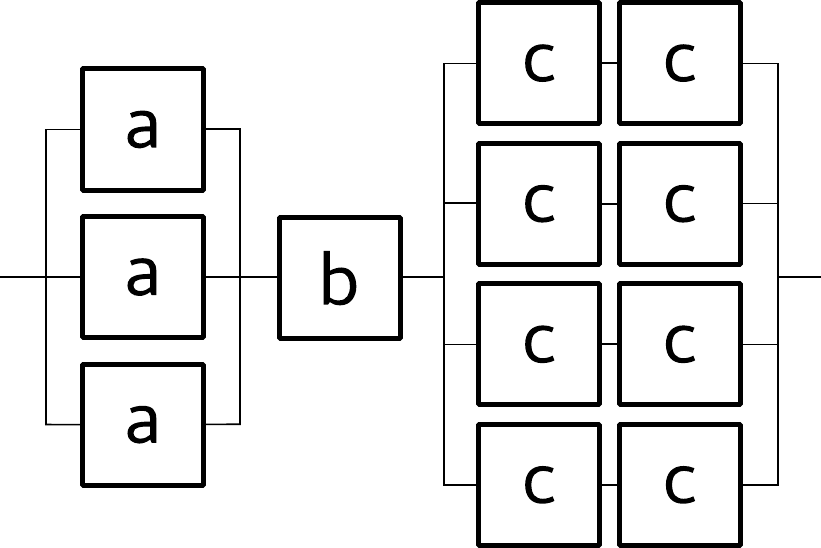}
    \caption{Schematic of a system composition consisting of three resource
        types: a,b,c, where the ratio from required to available is for a 1:3,
    for b 1:1, and for c 2:8.}
    \label{fig:organisation_model:metrics:system_decomposition}
\end{figure}

\begin{definition}[Functional reliability]
$R(\widehat{A},\mathcal{F})$ denotes the \textbf{reliability} of a set of required
functionalities $\mathcal{F}$ which is provided by an agent $\widehat{A}$.
\end{definition}
The computation of $R(\widehat{A},\mathcal{F})$ is based on the functional
decomposition of the agent type $\widehat{A}$ into atomic resources.
For each resource a redundancy at component level (cf.
\cite{Rausand:2009:ReliabilityTheory}) is assumed.
As a heuristic the redundancy is computed based on a type
partitioning considering all resources which have no further dependencies.
All resources of the same type are modeled as subsystems, which again form a
serial system.
Figure~\ref{fig:organisation_model:metrics:system_decomposition} illustrates
this modeling approach.

For each subsystem which is composed of a single resource type the redundancy is computed for $r$ required instances,
$n$ available instances and the probability of survival $p$ for the resource
type:

\begin{align}
    rsub(r,n,p) &= \begin{cases}
        1 - \left[1-p^{r}\right]^{ \frac{n}{r}} & \text{, where } n \geq r \\
        0 & \text{ otherwise} \\
    \end{cases}
\end{align}

The function $req$ maps a set of functionalities $\mathcal{F}$ to the required number of
instances for each resource type:
\begin{align}
    req(\mathcal{F})=\{req_1,\ldots,req_{|req(\mathcal{F})|} \} \quad,
\end{align}
where $req_i$ represents the minimum cardinality of a resource type $i$ to fulfill all functionalities in $\mathcal{F}$.

The function $avl$ maps an agent type to the number of maximum available
resources with respect to a functionality set $\mathcal{F}$.
Only resources that can contribute to the provision of $\mathcal{F}$ need to be
considered:
\begin{align}
    avl(\widehat{A},\mathcal{F})=\{avl_1,\ldots,avl_{|req(\mathcal{F})|}\} \quad,
\end{align}
where $avl_i$ represents the maximum cardinality of a resource type $i$
available in the general agent type $\widehat{A}$.

Resources lead to a heuristic system structure as shown in
Figure~\ref{fig:organisation_model:metrics:system_decomposition} using serial
and parallel systems.
Based on this structure, an agent's reliability with respect to a set of functionalities is defined as:
\begin{align}\label{equation:reliability:functionalities}
    R(\widehat{A},\mathcal{F}) = \prod_{i = 1}^{|req(\mathcal{F})|}
    rsub(req_i,avl_i,p_i) \quad,
\end{align}
where $p_i$ represents the probability of survival for a resource type $i$.

\subsubsection*{Example}
Before introducing this example note that functionalities can dependent upon other functionalities or resources. Hence, an availability of functionality $f_1$ might imply the availability of a functionality set $\mathcal{F}_1 = \{ f_1, f_2,f_3 \}$, where $f_2$ and $f_3$ are direct dependencies of $f_1$.
In this example a functionality \ontoKey{LocationImageProvider} depends on a functionality
\ontoKey{ImageProvider} and a functionality \ontoKey{MoveTo}.
The requirements for the \ontoKey{ImageProvider} are one of each resource: \ontoKey{Camera} and
\ontoKey{PowerSource}. \ontoKey{MoveTo} requires one of each resource: \ontoKey{Localization},
\ontoKey{Locomotion}, \ontoKey{Mapping}, \ontoKey{PowerSource}.

Table~\ref{tab:organisation_model:metrics:example} lists the cardinalities and probabilities of
survival for an atomic agent type \ontoKey{SherpaTT} with relevant resources.
\begin{table}
\centering
\caption{Resources of the agent type \ontoKey{SherpaTT} which are relevant to provision
    the functionality \ontoKey{LocationImageProvider}.}
\label{tab:organisation_model:metrics:example}
\small
\begin{2maztabular}{llcc}
    \textbf{Resource} & \textbf{$req_i$} & \textbf{$avl_i$} & \textbf{$p_i$}   \\\midrule
    Localization & 1 & 1 & 0.95 \\
    Locomotion   & 1 & 1 & 0.95 \\
    Mapping      & 1 & 1 & 0.95 \\
    PowerSource  & 1 & 1 & 0.95 \\
    Camera       & 1 & 2 & 0.95 \\
\end{2maztabular}
\end{table}
This agent type provides the functionality \ontoKey{LocationImageProvider} with
a redundant camera system, and otherwise a series system.
According to Equation~\ref{equation:reliability:functionalities} the probability
of survival for the functionality \ontoKey{LocationImageProvider} is:
\[ 
	P = (1 - (1-0.95)^2) \cdot 0.95^4 \approx 0.81
\]
A composite agent which has an additional atomic agent \ontoKey{PayloadBattery} can increase the redundancy of the resource
\ontoKey{PowerSource} by 1. This leads to an increase of the probability of survival for
the functionality \ontoKey{LocationImageProvider}, since now two redundant subsystems
exist:
\[
	P = (1 -(1-0.95)^2)^2 \cdot 0.95^3 \approx 0.85
\]

\subsubsection{Efficacy} 
Efficacy in the context of \gls{MoreOrg} describes the ability of a reconfigurable multi-robot system to provide a particular functionality.
To measure an agent's efficacy an objective has to be given, here as a set of required functionalities.
The identification of efficacy leads only to a binary result: either the
organization supports the given functionality or not.

% function and general agent
The definition of support can be viewed as an analysis of the redundancy level of components with respect to a required set of functionalities. 
The bottom-up definition of functionality support, eventually leads to the definition of an agent's efficacy:
\begin{align}
    \textit{efficacy}(\widehat{A}, \mathcal{F}) = \begin{cases}
        1 & \text{ } support(\widehat{A}, \mathcal{F}) \geq 1 \\
        0 & \text{ otherwise} \\
    \end{cases}
\end{align}

This definition of efficacy illustrates our approach to use resource summation to actively exploit redundancies for the creation of functional systems.
For efficacy only a sufficient component count is relevant, while the computation of the safety objective will take into account any excess resources.

\subsubsection{Operation Cost}
Typically, robotic systems consume electrical energy and \gls{MoreOrg}
additionally expects a definition of all agents nominal power consumption.
Since we do not assume a homogeneous set of robots and might also operate in varying coalitions agents will have a varying power consumption.
Therefore, operation time only is not an accurate cost measure.
Instead, \gls{MoreOrg} uses the total energy consumption of a reconfigurable multi-robot system as cost measure.

Power consumption can vary over time with the type of activity, but
\gls{MoreOrg} assumes a constant power consumption of all operative
atomic agents and leaves a more sophisticated estimation, e.g., based on a
functionality-based power consumption model, as future enhancement.
\gls{MoreOrg} estimates the operation cost as total consumed energy
for all agents based on the nominal power consumption:
\[
	ocost(\widehat{A},t) = \sum_{\hat{a} \in \widehat{A}} \gamma_{\widehat{A}}(\hat{a})(pw(\hat{a}) \cdot t)
\]

\subsection{Organization Properties}\label{sec:organisation_properties}
In contrast to atomic and composite agents, an organization property can differ in its structure, i.e., might change its coalition structures over time.
Therefore we describe here the coalition structure properties of a reconfigurable multi-robot system, as well as the cost to change between coalition structures.

\subsubsection{Coalition Structure Properties}
\paragraph{Goal Dependant Reliability}
The organizational structure intends to support activities
of its member agents that help to achieve and maximize the shared organizational goals.
Agents can operate in parallel at different physical locations.
The objective for an active coalition structure $CS = \{GA_{1},\ldots,GA_{|CS|}\}$ of an
organization is therefore described by a corresponding set of functionality sets denoted
$FS = \{ \mathcal{F}_1, \ldots \mathcal{F}_{|CS|} \}$, and 
$a2f: CS \rightarrow FS$ allows to map each operative agent $GA_{i}$ to the
required functionality set $\mathcal{F}_i$.
The current redundancy of the organization is then the minimum achievable level of
redundancy:
\begin{align}
    R(CS,FS) &= \operatorname*{min}_{A \in CS}  R(\widehat{A}, a2f(A))
\end{align}

Note that this computation is not used in the planning approach, but will be used in the future to identify coalition structures with critically unbalanced resource distribution.

\paragraph{Efficacy}
The efficacy of an organization's current coalition structure is computed from all operative agents' efficacy:
\begin{align}
    \textit{efficacy}(CS,\mathcal{F}) = \min_{A \in CS} \textit{efficacy}(\widehat{A},
    \mathcal{F})
\end{align}
Here, all operative agents in a coalition structure need to support the functionalities in $\mathcal{F}$.

Note, that \gls{MoreOrg} implements the optimal coalition structure generation algorithm developed by \textcite{Rahwan:2009:AnytimeCSF} to search for a coalition structure where $\textit{efficacy}(CS, \mathcal{F}) = 1$.
This optimisation approach is used, e.g., to validate the feasibility of a transition of a set of agents between two locations. Since only mobile agents can relocate, a coalition structure is required such that $\textit{efficacy}(CS,\{MoveTo\}) = 1$.

\subsubsection{Operation Cost}\label{sec:organisation_properties:operation_cost}
Reconfiguration of an organization contributes to operation cost, since transitions between coalition structures require time and energy.
The time to transition from one coalition structure $CS^A_i$ to another $CS^A_j$ when $CS^A_i \neq CS^A_j$ is therefore estimated with a heuristic function.
The heuristic firstly assumes basic cost for the number of atomic agents which
are involved in the reconfiguration.
Secondly, additional and significantly higher cost arise from the need to
coordinate multiple agents to exchange atomic agents or to merge.
Therefore, the reconfiguration cost function to form a single agent from an existing
coalition structure takes into account the number of general agents, equal to the partitions of the coalition structure $CS$, and the total number of involved atomic agents:
\begin{align}
    \rho(GA,CS) = t_a \cdot |CS| + t_b \cdot |GA| \quad,
\end{align}
where $t_a$ and $t_b$ are heuristic time constants.
Robotic experiments are required to
establish a realistic estimate for the magnitude of these parameters.
Our real world experiments give an indication for
these parameters for small teams, so that a default setting of $t_a=600\,s$ and $t_b=100\,s$ applies.
The values are estimates which consider time for additional error handling.
The overall reconfiguration cost to transition from a coalition structure $CS^A_i$ to
another $CS^A_j$ is defined as:
\begin{align}
    \rho(CS^A_i,CS^A_j) = \sum_{GA \in CS^A_j} \rho(GA,CS^A_i)
\end{align}
Note that the reconfiguration cost heuristic does not account for
relocation cost. Instead we assume that all agents taking part in a reconfiguration process operate in direct proximity.
Thus, this heuristic penalizes an involvement of an increasing number of agents
to extract a new one.  The first term penalizes the total number of involved
independent agents to form a new agent with; each additional agent increases
communication and coordination cost, as well as the likelihood for failures.
The second term accounts for the fact, that only a selected set of atomic agents is involved in a reconfiguration - the fewer the better.

\section{Exploitation of Redundancies}\label{sec:planning}
To exploit reconfigurable multi-robot systems we use a constraint-based mission planning approach.
Constraints allow to define the essential characteristics of a potential solution, by letting a user specify task requirements.
The collected constraints then define a partial-ordered plan for which a suitable full solution has to be found by translating the mission requirements to an agent allocation and constellation problem.
Agent allocation here means, that the exact organization structure and its development over time throughout a mission is initially unknown.
What is known apart from the given constraints, however, are the total available atomic agents, their associated resources and the model to combine atomic agents to composite agents in order to support required functionalities.

We implemented the planner \gls{Templ}~\cite{Roehr:2018:IBERAMIA} to address the constraint-based planning problem for reconfigurable multi-robot systems.
This planner uses \gls{MoreOrg} as organization model, but adds a temporal and spatial dimension to problem definitions.

\begin{definition}[\emph{\textbf{Spatio-temporal
    requirement}}]\label{definition:stqe}
A spatio-temporal requirement is represented as a spatio-temporally qualified expression $s$, which
describes the functional requirements and agent instance
requirements for a time-interval and a location: 
\[
    s = (\mathcal{F},\widehat{A}_s)@(l,[t_s,t_e]) \text{,}
\]
where $\mathcal{F}$ is a set of
functionality constants, $\widehat{A}_s$ is the general agent type representing the
required agent type cardinalities, $l
\in L$ is a location variable, and $t_s,t_e \in T$ are temporal variables
describing a temporal interval with the implicit constraint $t_s < t_e$.
\end{definition}

Each spatio-temporal requirement represents a persistence constraint, i.e., the
requirements have to hold throughout the time interval.
The mission specification allows to relate spatio-temporal requirements
to the organization model which defines agent types and
functionalities along with the associated properties.
\begin{definition}[\emph{\textbf{Mission}}]
    A robotic mission is a tuple $\mathcal{M} =
    \langle \widehat{A},STR,\mathcal{X},\mathcal{OM}, T, L \rangle$, where the
    general agent type $\widehat{A}$
describes the available agent types, $STR$ is a set of
spatio-temp\-orally qualified expressions, $\mathcal{X}$ is a set of
constraints, 
$\mathcal{OM}$ represents the organization model (here \gls{MoreOrg} as described in Section~\ref{sec:organisation_modelling}), T is the set of timepoints
and L the set of locations.
\end{definition}
\subsection{Mission Constraints}
Constraints in $\mathcal{X}$ can refer to spatio-temporally qualified expressions and the initial state of a
mission is defined by the earliest timepoint and binds available agents to their
starting depot.
The earliest timepoint is $t_0\in T$ and $\forall t \in T, t\neq t_0: t > t_0$.
Note that this planning approach neither requires a single starting location for all agents nor a single final destination. An example for a mission is given in Table~\ref{tab:evaluation:mission}.
\begin{table*}%[ht!]
\centering
\caption{Temporal constraints for a mission 
    $\mathcal{M} = \langle \widehat{A},STR,\mathcal{X},\mathcal{OM},T,L
\rangle$.
}
\label{tab:planning:temporal_constraints}
\footnotesize
\begin{2maztabular}{llp{8cm}}
\textbf{Name}  & \textbf{Syntax} & \textbf{Description} \\\midrule
\textbf{temporal relation} & $\langle t_n,REL,t_m\rangle$ & $t_n$ and $t_m$ are qualitative
timepoints and $REL$ is the set of permitted relations, so that $REL \subseteq \{<,>,=\}$~\parencite{Dechter:2003:TCN} \\
%\item[\textbf{unification}] 
\textbf{min duration} & $minDuration(t_n,t_m,d)$ & sets a lower bound for the
duration of a time interval: $t_n - t_m \geq d$, where $t_n$ and $t_m$ are two
qualitative timepoints $d \in \mathbb{R}^{+}$;
implies the qualitative relationship $t_n > t_m$ \\
\textbf{max duration} & $maxDuration(t_n,t_m,d)$ & sets an upper bound for the duration
of a time interval: $t_n - t_m \leq d$, where $t_n$ and $t_m$ are two
qualitative timepoints $d \in \mathbb{R}^{+}$; implies the qualitative relationship $t_n > t_m$
\\
\end{2maztabular}
\end{table*}
\paragraph{Temporal Constraints}
Temporal constraints are listed in Table~\ref{tab:planning:temporal_constraints}. They allow to defined the temporal relation between spatio-temporal constraints in a relative and absolute manner.
\paragraph{Model Constraints}
\gls{Templ} implements a subset of feasible (meta-)constraints
(see Table~\ref{tab:planning:model_constraints}).
Model constraints set requirements for agent types and agent roles.
They allow bounding the cardinality of agent types so that the combinatorial
search problem can be limited according to a least-commitment principle. 
Equality constraints allow to restrict
agent routes partially or even completely.
Requiring the minimum equality of a
single agent type over the full mission defines the
full route for a single atomic agent of this type.
Thereby, modelling constraints allow to detail a mission.
In general, constraints apply to the
dimensions space, time, agent types, and roles.
%
%The equality constraint allows to define partial as well as complete routes for
%agents, since the constraint allows to enforce that a designed agent
%combination, e.g., fulfilling a functionality, has as to be
%present for.
%
\begin{table*}%[ht!]
\centering
\caption{Model constraints, where $S \subseteq STR$ and
        $\widehat{A}_{s}$ represents the general agent type requirement of $s
    \in S$.
}
\label{tab:planning:model_constraints}
\footnotesize
\begin{2maztabular}{llp{8cm}}
\textbf{Name}  & \textbf{Syntax} & \textbf{Description} \\\midrule
\textbf{min cardinality} & $minCard(S,\hat{a},c)$ & 
    Minimum cardinality constraint $\forall s \in S : \gamma_{\widehat{A}_s}(\hat{a}) \geq
    c$, where $c \geq 0$ \\
\textbf{max cardinality} & $maxCard(S,\hat{a},c)$ &
    maximum cardinality constraint corresponding to $minCard$ so that
    $\forall s \in S: \gamma_{\widehat{A}_s} \leq c$, where $c \geq 0$ \\
\textbf{all distinct} & $allDistinct(S,\hat{a})$ &
    $\forall s \in S: \bigcap A_s^{\hat{a}} = \emptyset$ \\
\textbf{min distinct} & $minDistinct(S,\hat{a},n)$ & 
    $\forall s_i,s_j \in S, i \neq j: \left| |A_{s_i}^{\hat{a}}| -
|A_{s_j}^{\hat{a}}| \right| \geq n$, where $n > 0$ \\
\textbf{max distinct} & $maxDistinct(S,\hat{a},n)$ &
    the equivalent maximum constraint to $minDistinct$, so that
    $\forall s_i,s_j \in S, i \neq j: \left| |A_{s_i}^{\hat{a}}| -
|A_{s_j}^{\hat{a}}| \right| \leq n$, where $n \geq 0$ \\
\textbf{min equal} & $minEqual(S,A_r)$ &
    %minimum existence of the same agent roles so that $\exists A_r \forall s \in
    %S: A_r \subset r(A_s)$, where $A_r \subset r(A)$, $A$ is the
    %available agent pool for a mission, and $A_s$ is the agent pool that
    %fulfils $s \in S$ \\
    minimum existence of the same agent roles so that $A_{eq} = \bigcap_{s \in
S} r(A_s)$ and $A_r \subset A_{eq}$, where $A_r \subset r(A)$, $A$ is the
    available agent pool for a mission, and $A_s$ is the agent pool that
    fulfils $s \in S$ \\
\textbf{max equal} & $maxEqual(S,A_r)$ &
    maximum existence of the same agent roles so that $A_{eq} = \bigcap_{s \in
S} r(A_s)$ and $A_{eq} \subset A_r$, where $A_r \subset r(A)$, $A$ is the
    available agent pool for a mission, and $A_s$ is the agent pool that
    fulfils $s \in S$ \\
\textbf{all equal} & $allEqual(S,A_r)$ &
    the constraint conjunction: $minEqual(S,A_r) \land maxEqual(S,A_r)$ \\
\end{2maztabular}
\end{table*}
\paragraph{Functionality \& Property Constraints}
Agents either comprise a functionality or they do not.
In effect, functionality is requested with a maximum cardinality of one,
which makes the introduction of a maximum function constraint unnecessary.
Note that this is a limitation of the current modelling approach.
However, the property of a agent providing a particular functionality can be of
importance, and the use of property constraints allows to narrow applicable
agents.
\begin{table*}
\centering
\caption{Functionality and property constraints.}
\label{tab:planning:functionality_constraints}
\footnotesize
\begin{2maztabular}{llp{8cm}}
\textbf{Name}  & \textbf{Syntax} & \textbf{Description} \\\midrule
\textbf{min function} & $minFunc(s,f)$ & 
functionality $f$ to be available at \gls{str} $s \in STR$, so that $f \in
\mathcal{F}^s$, where $\mathcal{F}$ represents the functionality requirements
associated with $s$ \\
    % with a minimal redundancy
%\item[\textbf{max-function}]
\textbf{min property} & $minProp(s,f,p,n)$ & 
constrain the numeric property $p_f$ of a functionality $f$ to be $p_f \geq n$,
where $n \in \mathbb{R}$ and $minProp(s,f,p,n)$ implies that $minFunc(s,f)$
holds\\
\textbf{max property} & $maxProp(s,f,p,n)$ &
equivalent maximum property value constraint to $minProp(s,f,p,n)$, so that
property $p_f \leq n$, $n \in \mathbb{R}$ \\
\end{2maztabular}
\end{table*}

\subsection{Planning and Optimization}
The (high-level) optimization problem for a given mission $\mathcal{M}$ can be stated as:
\begin{eqnarray*}
    \operatorname*{minimize}_{\mathcal{M}^{*}}&& cost(\mathcal{M}^{*}, \mathcal{M}) \\
    \text{subject to} && STR \text{ and } \mathcal{X}
\end{eqnarray*}
, where
\begin{eqnarray*}\label{chapter:planning:eqn:cost_function}
    STR && \text{spatio-temporal requirements} \\
    \mathcal{X} && \text{mission constraints} \\
    \mathcal{M}^{*} && \text{solution to mission }\mathcal{M} \\
    cost(\mathcal{M}^{*},\mathcal{M}) &=& \alpha E(\mathcal{M}^*) \\
                                      &+& \beta SAT(\mathcal{M}^{*},\mathcal{M}) \\
                                      &+& \epsilon SAF(\mathcal{M}^{*}, \mathcal{M})
\end{eqnarray*}
The multi-objective cost function is based on the
heuristics listed in Table~\ref{tab:planning:heuristics}, and can be evaluated as soon as a solution for a mission exists.

The three objectives are presented in the cost function:
efficiency through the energy cost function, efficacy through checking the level
of fulfillment, and safety as a redundancy dependent survival metric; for
balancing the parameters $\alpha$, $\beta$ and $\epsilon$ can be used.
Note that safety and fulfillment have a value range $[0,1]$, so
that $\alpha$ should account for normalization to $[0,1]$ for the energy cost;
$\alpha$ therefore comprises a factor $E_{max}^{-1}$, where $E_{max}$ is the maximum energy cost which is either the allowed one or is extracted from existing mission solutions. The latter is done for our evaluation in Section~\ref{sec:evaluation}.

The importance of operation cost is controlled via $\alpha$, a higher value will
lead a preference of missions which require less energy.
The time of an agent's operation depends upon estimated travel cost, time for
the actual requested operation or task, and time for reconfiguration.
Our approach extends an approach used by Wurm et al.~\cite{Wurm:2013:Coordination} to estimate the cost based on the travel time between two locations.
We suggest to use nominal speed $v_{nom}$ as default property for atomic agents,
so that based on this information a duration estimate for location changes of mobile agents can be computed.
As long as no better estimation or other routing constraints are available, the line-of-sight distance between
two locations is still the basis for the cost computation.

Parameter $\beta$ controls the penalty for missions that can only be partially
fulfilled, here the heuristic assumes that each requirement is of equal
importance. Future approaches should also take a priority into account.

The preference of safer operations and thus agents with higher redundancy is
control via $\epsilon$.
In principle, a high negative value of $\epsilon$ leads to a preference for a solution with a single, yet highly redundant
agent that can solve the mission.

For the search for an optimal transition between coalition structures it has to be considered, that due to a high degree of redundancy a safer coalition structure might lead to lower efficiency.
Therefore, any optimization has to trade safety and efficiency against each
other and can lead only to a pareto-optimal solution.

\subsection{Algorithm}
To tackle the given optimization problem one or more solutions for a given mission have to be
identified.
Due to the need for various possible mappings between required functionality
and satisfying agents, the problem cannot be directly solved in a classical
form, e.g., with Linear Programming.
The problem needs to be transformed with the help of the
\gls{MoreOrg}, so that existing optimization approaches can be combined to
perform local optimization, here
combinatorial optimization, coalition structure optimization and min-cost flow
multi-commodity flow via Linear Programming.

\begin{figure}
\includegraphics[width=\columnwidth]{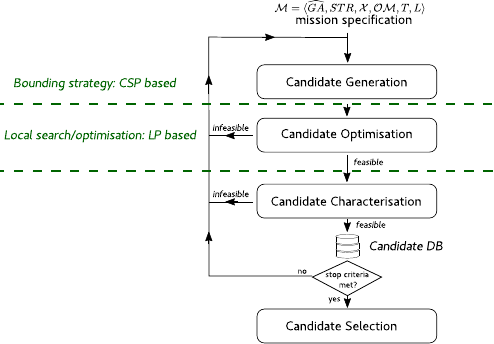}
\caption{Planning with \gls{TemPl} uses a constraint-based solution candidate generation, where each candidate is characterized according to the mission objectives.}
\label{fig:planning:templ}
\end{figure}

The basic algorithm that we are currently using is illustrated in Figure~\ref{fig:planning:templ}.
We will outline the general approach and refer to \cite{Roehr:2018:IBERAMIA} for
further details.
The algorithm involves a generation candidate generation, which accounts for major constraints but not all required ones. 
Only after candidate optimization and subsequent characterization it will be known whether the suggested solution candidate is feasible, i.e., whether the resulting quantitative temporal network is consistent and whether required reconfigurations and location transitions can be performed.
Since we also permit partial solutions, i.e. efficacy $< 1$, feasibility does not necessarily imply that all spatio-temporal requirements of a mission are satisfied, but only that the activities that are part of the suggested solution can be performed.
Generating solution candidates with an efficacy $< 1$
\begin{inparaenum}[(a)]
\item accounts for a future prioritization of requirements, and
\item permits the identification of pareto-optimal solutions with respect to safety, efficacy and efficiency.
\end{inparaenum}
\begin{table*}
    \caption{Heuristic cost computation on the solution $\mathcal{M}^{*}$ for
    a mission $\mathcal{M}$ (adapted from \cite{Roehr:2018:IBERAMIA})}
\label{tab:planning:heuristics}
\centering
\footnotesize
\begin{2maztabular}{p{3cm}lp{8cm}}
\textbf{Name} & \textbf{Syntax} & \textbf{Description} \\\midrule
\textbf{distance} & $d(a, \mathcal{M}^{*})$ & traveled distance of an agent $a$ in $\mathcal{M}^{*}$ \\
\textbf{operation time} & $op(a, \mathcal{M}^{*})$ & time
horizon of the mission; any location change introduces a lower bound $\Delta t_{min}$ for time
intervals by assuming a traversal with the mobile agent's nominal velocity
$v_{nom}(\hat{a})$, i.e., $\Delta t_{min} \geq \frac{d(a,\mathcal{M}^{*})}{v_{nom}(\hat{a})}$ \\
\textbf{energy} & $E(\mathcal{M}^{*})$ & $E(\mathcal{M}^{*}) = \sum_{a \in A} ocost({\hat{a}}, op(a,\mathcal{M}^{*}))$ as overall consumed energy per mission, by summing the consumed energy per agent $a$ to perform $\mathcal{M}^{*}$; \\
\textbf{safety} & $SAF(\mathcal{M}^*,\mathcal{M})$ &
    $SAF(\mathcal{M}^{*},\mathcal{M}) = \min_{s \in STR} R(A^{*}_s, \mathcal{F}_s)$, where R denotes the functional reliability function (see Section~\ref{sec:organisation_properties:safety})
    , $\mathcal{F}_s$ is the required functionality to satisfy $s$ and $A^{*}_s$ the available and assigned agent in mission solution $\mathcal{M}^{*}$.\\
\textbf{fulfilment} & $SAT(\mathcal{M}^{*},\mathcal{M})$ & 
        $SAT(\mathcal{M}^{*},\mathcal{M}) = \frac{1}{|STR|}\sum_{s \in STR} sat(s, \mathcal{M}^{*})$
        represents the ratio of fulfilled requirements, where 
\[ 
sat(s, \mathcal{M}^{*})= 
\begin{cases}
 0 & \text{if constraint \emph{s} is not satisfied in } \mathcal{M}^{*} \\
 1 & \text{if constraint \emph{s} is satisfied in } \mathcal{M}^{*} \\
\end{cases}
\]\\%Weighting of requirements allows to define priorities to the requirements.
\end{2maztabular}
\end{table*}

Initially the planner aims at generating a resource efficient solution, by reducing the number of involved agents. It does so by bounding the number of agents that are assigned to mutually-exclusive requirements through two parameters: $\psi_m$ and $\psi_{\lnot m}$.
The default is $\psi_m = 0$ and $\psi_{\lnot m} = 0$, which means, that a set of mutually-exclusive requirements only get the minimum needed resources to resolve their conflict: the parameters $\psi_m$ and $\psi_{\lnot m}$ refer to the number of additionally permitted mobile and immobile agents respectively to be assigned to mutually-exclusive requirements. In effect, these parameters can increase the resource usage and level of redundancy for a solution, and at the same time increase the options to find a solution that satisfies all requirements.
A resulting increase of the number of agents can come, however, at a price:
\begin{inparaenum}[(a)]
\item higher complexity of the problem since additional agents allow for exponentially more reconfiguration options, and
\item higher operational cost, i.e., lower efficiency.
\end{inparaenum}

\section{Evaluation}\label{sec:evaluation}
\begin{figure*}
	\centering
	\includegraphics[width=\textwidth]{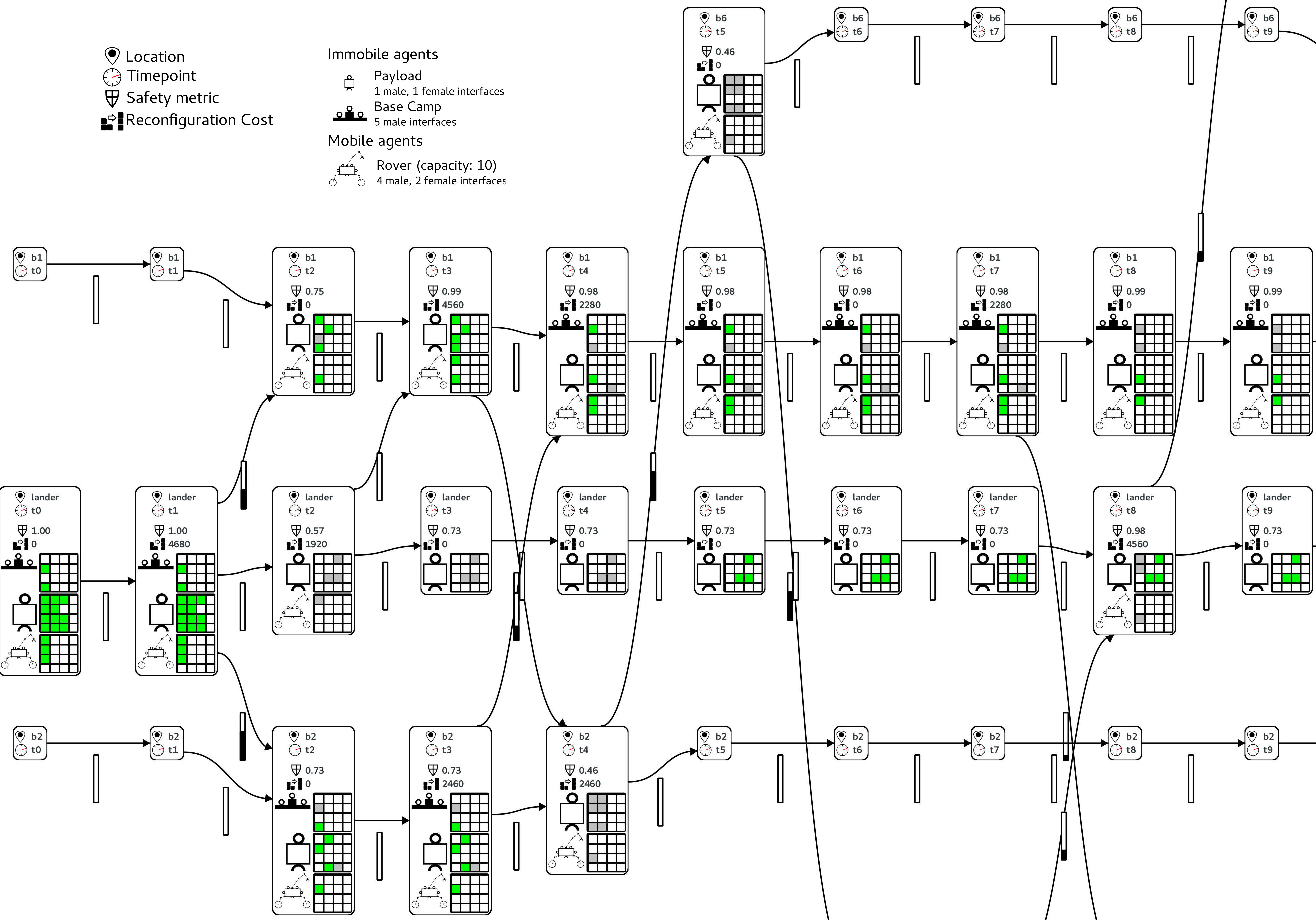}
	\caption{Identified solution for the exemplary space mission after 20\,min of search ($\psi_m=2, \psi_{\lnot m} = 0$) (to maintain readability only a part of the graph is placed here). Bars are annotating edges to illustrate transport capacity consumption, a green square identifies a required and available agent, while a grey square identifies an available though not required agent. Each constellation is attributed with the safety metric (probability of survival) and reconfiguration cost (in s).}
	\label{fig:planning:templ:mission_complex:space_scenario}
\end{figure*}

For an experimental evaluation of the planning approach we look at a simulated space mission, which has been outlined by \textcite{Sonsalla:2014:TransTerra}.
The major goal of the mission is to place scientific payloads in a lunar environment at a predefined set of locations for science goals.
These target science goals are cast into a respective mission specification which is listed in Table~\ref{tab:evaluation:mission}. 
Note that each location is defined by longitude and latitude in the specification;
\gls{TemPl} uses Mercator projection to convert these into Cartesian coordinates.
A corresponding subsection of a solution found by \gls{TemPl} is shown in Figure~\ref{fig:planning:templ:mission_complex:space_scenario}.
With respect to the safety measure the starting assignment is ignored by assuming a probability of survival of 1.0. This is done to avoid an initial bias of the safety objective.
To compute the safety objective only relevant spatio-temporal requirements are analyzed. Hence, agents which are available at a space time point, but are not actually required there (represented with gray colored boxes in Figure~\ref{fig:planning:templ:mission_complex:space_scenario}) do not affect the safety objective (see for instance location $b6$, timepoint $t5$).

\begin{figure*}
    \centering
    \includegraphics[width=0.8\textwidth]{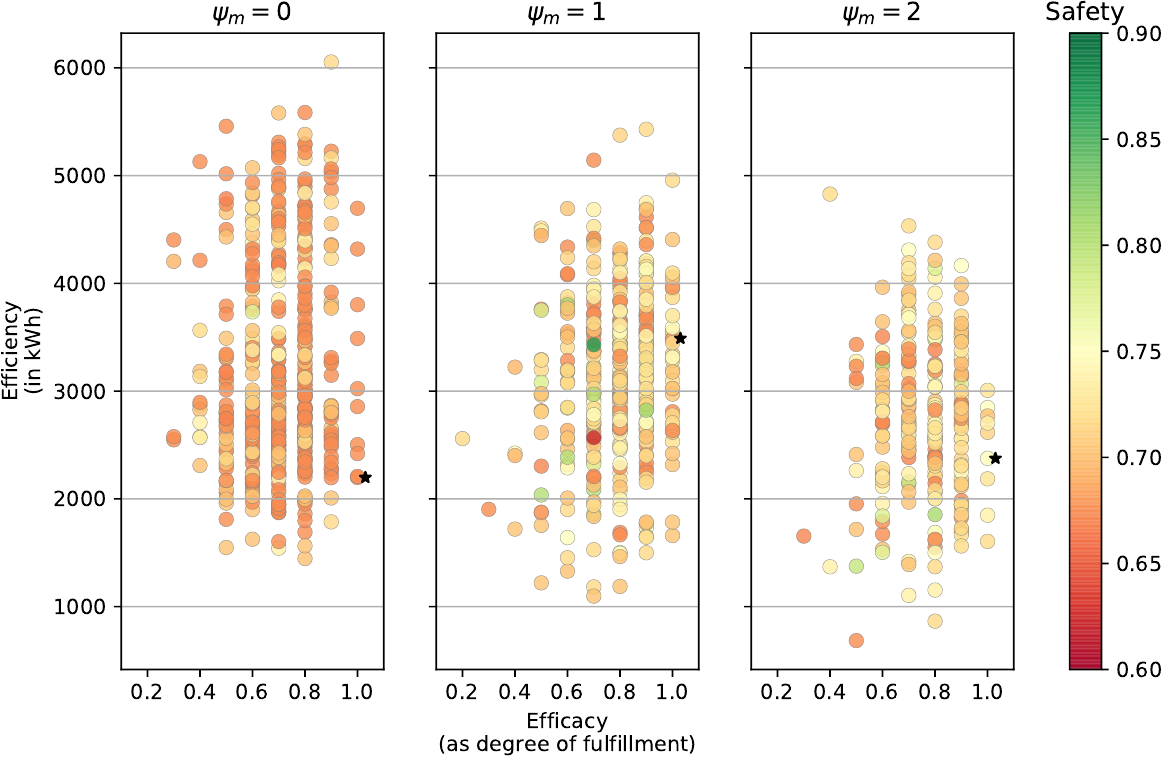}
    \caption{Solution landscape compared for different bound settings. Black
    star-shaped markers identify the current local best solutions where the
    cost function is parameterized with $\alpha=1.0$, $\beta=-100.0, \epsilon=-10.0$. The results for each setting of $\psi_m$ (efficiency, safety) are:
    $\psi_m=0$: (2198.25\,kWh, 0.674), $\psi_m=1$: (3489.5\,kWh, 0.754), $\psi_m = 2$: (2376.11\,kWh, 0.747)
}
    \label{fig:planning:templ:evaluation:solution_analysis}
\end{figure*}

\begin{figure*}
    \centering
    \subfloat[Analysing efficiency with respect to efficacy.\label{fig:planning:templ:evaluation:solution_analysis:efficiency}]{
        \includegraphics[width=.8\textwidth]{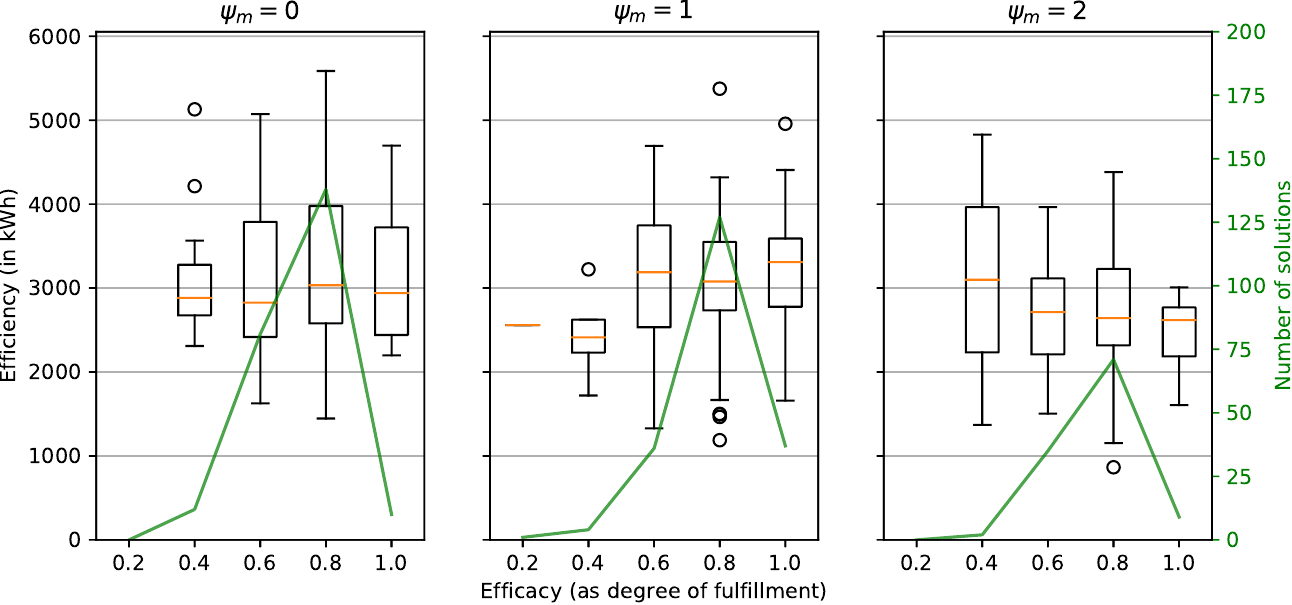}}\\       
	\subfloat[Analysing safety with respect to efficacy.\label{fig:planning:templ:evaluation:solution_analysis:safety}]
	{\includegraphics[width=.8\textwidth]{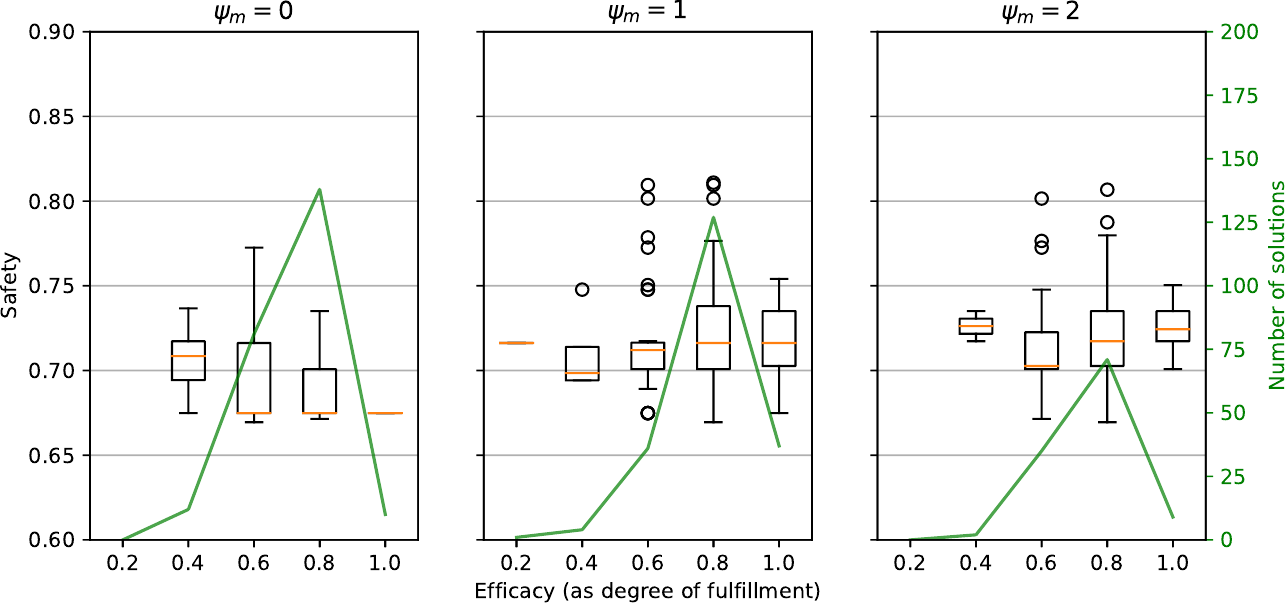}}\\
    \subfloat[Efficiency with respect to safety, where efficacy = 1.0.\label{fig:planning:templ:evaluation:solution_analysis:pareto}]
	{\includegraphics[width=.8\textwidth]{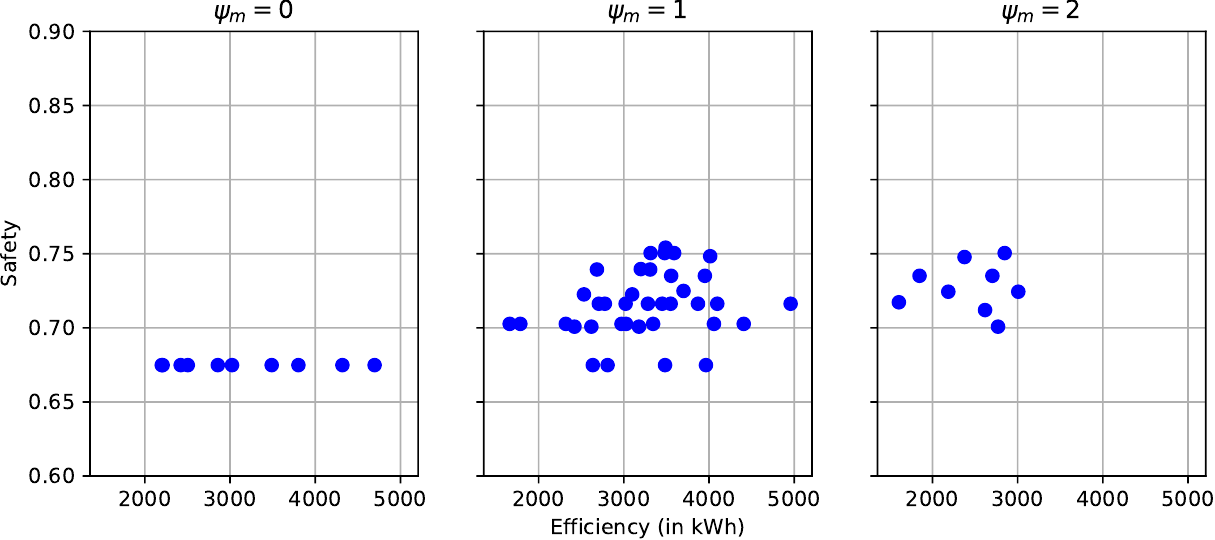}}
	
    \caption{Analysing the solution landscape of the space mission
    with respect to the optimization objectives.}
    \label{fig:planning:templ:evaluation:solution_analysis:objectives}
\end{figure*}

\begin{table*}
\caption{An exemplary outline of a space exploration mission for a team of reconfigurable robots.}
\label{tab:evaluation:mission}
\begin{footnotesize}
\begin{flalign*}
    \widehat{GA}  =& \{(BaseCamp,5),(CREX,2),(CoyoteIII,3),(Payload,16),(SherpaTT,3)\} \\
    STR       =& \bigl
    \{(\{(BaseCamp,5),(CREX,2),(CoyoteIII,3),(Payload,16),(SherpaTT,3)\})@(lander,[t_{0},t_{1}]), \,\\
     & (\emptyset,\{(Payload,3)\})@(lander,[t_{5},t_{10}]), \\
     & (\{LocationImageProvider,EmiPowerProvider\},\{(Payload,3)\})@(b_{1},[t_{2},t_{3}]), \,\\
     & (\emptyset,\{(Payload,1)\})@(b_{1},[t_{3},t_{14}]), \, \\
     & (\{LogisticHubProvider,LocationImageProvider,EmiPowerProvider\},\{(Payload,3)\})@(b_{2},[t_{2},t_{3}]),\, \\
     & (\emptyset,\{(BaseCamp,1)\})@(b_{1},[t_{4},t_{7}]),\,
     (\{LocationImageProvider\},\{(Payload,3)\})@(b_{4},[t_{6},t_{7}]),\, \\
     & (\emptyset,\{(Payload,3)\})@(b_{4},[t_{8},t_{9}]),\, (\emptyset,\{(Payload,1)\})@(b_{6},[t_{10},t_{14}]),\, (\emptyset,\{(Payload,3)\})@(b_{7},[t_{12},t_{14}])\, \bigr \}\\
   \mathcal{X}  =& \{ t_0 < t_1, \ldots, t_{13} < t_{14} \} \\
   \mathcal{OM} =& \{ \lnot mobile(BaseCamp), mobile(CREX), tcap(CREX) = 2, mobile(CoyoteIII), tcap(CoyoteIII) = 4, \\
                 & \lnot mobile(Payload), mobile(SherpaTT), tcap(SherpaTT) = 10, \ldots \} \\
    T =& \{ t_0, \ldots, t_{14} \} \\
    L =& \{ lander = (lat: -83.82009, long: 87.53932, moon), b_{1} = (lat: -84.1812,  long: 87.60494, moon), \\
      &  b_{2} = (lat: -83.96893, long: 86.75471, moon), b_{3} = (lat: -83.66856, long: 87.42557, moon), \\
      &  b_{4} = (lat: -83.54570, long: 87.09851, moon), b_{5} = (lat: -83.82009, long: 84.66000, moon), \\
      &  b_{6} = (lat: -83.77371, long: 84.70960, moon), b_{7} = (lat: -83.34083, long: 84.64467, moon)\,  \}
\end{flalign*}
\end{footnotesize}
\end{table*}
%\caption{Mission specification for an anticipated space mission.
%Each location is defined by longitude and latitude;
%\gls{Templ} uses Mercator projection~\parencite{PROJ:2018:Mercator} to
%convert these coordinates.
%}
%\label{fig:planning:templ:evaluation:space_mission:spec}
%\end{figure}
\gls{TemPl} has been used to search for solutions to this mission scenario
with a setting of $\psi_{\lnot m}=0$ and $\psi_m$ in the range of 0 to 2.
The search has been split into epochs with a maximum allowed planning time of
60\,s. After 60\,s search has been reinitialized in order to escape from local
minima. 
Planning has been stopped either after memory
has been exhausted  or when the total planning time of 20\,min was exceeded (Intel i7-4600Um 12\,GB RAM).
A higher setting of $\psi_m$ requires additional agents to be used for the
planning approach, so a higher computation time per solution is to be expected.
Note that while the setting with $\psi_m=0$ resulted in a stable range below 4\,s per solution
candidate, the required computation time per solution increases for $\psi_m =2$ to up to 12\,s - in combination with the total planning time this explains the lower number of solutions found for higher settings of $\psi_m$.
%The \gls{LP} problem sizes remain, however, at a comparable level, despite the
The comparison of solutions based on efficacy, efficiency and safety
is shown in Figure~\ref{fig:planning:templ:evaluation:solution_analysis} with additional details provided in Figure~\ref{fig:planning:templ:evaluation:solution_analysis:objectives}.

%bound relaxation as shown in Figure~\ref{fig:planning:templ:evaluation:lpstats}.
%

What can be seen is that the expected increase in redundancy, due to the higher setting of the parameter $\psi_m$ leads to solution candidates with a consistently higher safety objective. This shows that $\psi_m$ is an effective control parameter.
By using an exemplary trade off setting for the cost function with $\alpha = -1$, $\beta = 100.0$, and $\epsilon = 10.0$ we can extract the solutions characteristics.
The black star-shaped marks indicate the best solutions according to the weights of the objective function.
While safety can be increased from below 0.7 to approx. 0.75 for solutions with efficacy of 1, the efficiency deteriorates, although only slightly for $\psi_m=2$ compared to $\psi_m=0$.

This analysis shows only an exemplary solution landscape, but at the same time the working of our modeling and planning approach for reconfigurable multi-robot system capabilities.
Our evaluation confirms that by increasing the options for agent assignments and at the same time the level of redundancy of the reconfigurable system, improved solutions can be found for application scenarios where safety might be an issue.

\section{Conclusion}\label{sec:discussion}
This paper has outlined a modeling approach for reconfigurable multi-robot systems, which permits an active exploitation of redundancies in multi-robot systems through physical reconfiguration.
The current approach does only moderately influence the level of redundancy in systems that are taking part in a robotic mission. This is mainly due to the complexity of the planning problem as result of exponentially many reconfiguration options. It led us to use several heuristics and initially target resource efficient solutions.
%We use the organization model to map from simple component structure to the full organization state to achieve the ability for dynamic agent assignments, so that inherent system redundancies can actively be exploited.
Clearly, this model might still not lead to solutions which are sufficiently good from a safety perspective, but it offers a basis for an automated use of reconfigurable multi-robot systems and an optimization strategy for organizational safety properties.
Our deliberate planning approach can give an advantage compared to reactive reorganization strategies found in swarm-like systems by avoiding dead-end configurations. Meanwhile a hybrid approach could be foreseen in a real application.
Furthermore, we plan to augment already found solutions by explicitly routing available or rather unused agents along the critical path of a mission.
This can be done without replanning until the known transport lines, which are established through mobile agents, are exhausted. It comes, however, again at the cost of efficiency.

The existing heuristics are based on assumptions and a limited set of
practical experiments.
As such they can clearly be interpreted as a weakness
of the modelling approach.
The given heuristics shall, however, serve as basis and examples to
develop better approaches to with reconfigurable multi-robot systems.
As such they firstly point to the need / benefit for a submodel, and
secondly act as placeholders.
All submodels and heuristics are subject of continued improvements and
efforts to detail the model further.
Probability of survival, for instance, as it is used and presented in this paper acts as a placeholder for
more sophisticated metrics, e.g., some which are based on
extensive empirical studies and specifications of subsystems.
Furthermore, we plan to define safety not only through spatio-temporal requirements, but instead adapt the model to reflect the risks resulting from reconfiguration, idling and relocation.
Since a solution is characterized after all agents have been assigned, a dedicated models, e.g., for component degradation and reconfiguration errors can be taken in account to improve the safety measure.
We are also interested in using graph and network analysis techniques such as percolation \cite{Newman:2010:Networks} in combination with replanning to test the options to respond to system failures.

Reconfigurable multi-robot systems combine the benefits of modular robots with capable robotic systems, and
we see significant potential in developing further strategies and planning approaches.
An automated exploitation will give designers of robotic mission a new degree of freedom.
Still, significant practical challenges remain to exploit reconfiguration with real robots: increasing the reliability of all involved atomic agents is one challenge, and establishing reliable reconfiguration maneuvers is another.
We do, however, outline here a feasible approach towards modeling and multi-objective planning for such systems.

% or


%\appendices

%\input{sections/appendix}
%% you can choose not to have a title for an appendix
%% if you want by leaving the argument blank
%\section{}

\section*{Acknowledgment}
This work has been supported by the German Space Agency (DLR Agentur) with
federal funds of the Federal Ministry of Economic Affairs and Energy for the
project TransTerrA under grant agreement 50RA1301 to implement the initial
planning approach. The continued evaluation and application in the project
Q-Rock has been supported by the Federal Ministry of Education and Research
under grant agreement 01IW18003. 
The continued development has been supported by
the Federal Ministry for Economic Affairs and Energy under grant agreement
50RA1701.
The author would like to thank all contributors who helped to realize the multi-robot team in the project TransTerrA.

\ifCLASSOPTIONcaptionsoff
  \newpage
\fi

%\IEEEtriggeratref{27}


%\bibliographystyle{IEEEtran}
%\bibliography{IEEEabrv,references}
%\bibliography{IEEEabrv,library}

\balance
\printbibliography
%\IEEEtriggercmd{\enlargethispage{-12in}}
%\begin{thebibliography}{1}
%\end{thebibliography}


%\newpage


%\vfill


\end{document}